\documentclass[11pt,a4paper]{article}
\usepackage[hyperref]{emnlp-ijcnlp-2019}
\usepackage{breakurl}
\usepackage{times}
\usepackage{latexsym}

\usepackage{booktabs}

\usepackage{url}

\usepackage{amsmath}
\usepackage{multirow}
\usepackage{adjustbox}

\usepackage{graphicx}  %Required
\usepackage{amsfonts}
\usepackage{subcaption}
\renewcommand{\vec}[1]{\mathbf{#1}} %make vector symbol bold

\usepackage{footnote}
\usepackage{epstopdf}
\usepackage{algorithm} %algorithm package
\usepackage{algpseudocode} %algorithm
\usepackage{stmaryrd}
\usepackage{booktabs,arydshln}

\usepackage{makecell}
\usepackage{color}
\usepackage{tikz}
\usepackage{pgf}
\usepackage{pgfplots}
\usepackage{tikz-qtree}
\usetikzlibrary{arrows,decorations.pathmorphing,backgrounds,positioning,fit,petri,shapes.misc, arrows.meta,shapes.geometric,decorations.markings,calc,shadows.blur,decorations.pathreplacing,quotes,matrix,shapes.symbols}
\definecolor{fontgray}{RGB}{44, 62, 80}
\definecolor{myred}{RGB}{235, 47, 6} %rgb()
\definecolor{summertime}{RGB}{245, 205, 121}
\definecolor{darkgrass}{RGB}{0, 148, 50}
\definecolor{myblue}{RGB}{0, 168, 255}
\definecolor{mygray}{RGB}{158, 158, 158}
\definecolor{puffin}{RGB}{250, 152, 58}
\definecolor{lowpurple}{RGB}{210, 180, 222}
\definecolor{lowblue}{RGB}{102,178,255}
\definecolor{lowred}{RGB}{245, 183, 177}
\definecolor{darkblue}{rgb}{0, 0, 0.5}

\def\pnode [#1]#2{
	\node[regular polygon,regular polygon sides=4, minimum size=1pt,fill=gray,#1, inner sep = 1.2pt] (#2) {};
} 
\tikzset{middlefactor/.style={decoration={
			markings,
			mark= at position #1 with {\pnode[]{}} 
		},postaction={decorate}},
	middlefactor/.default=0.5
}

\newcommand{\squishlist}{
	\begin{list}{$\bullet$}
		{ \setlength{\itemsep}{0pt}
			\setlength{\parsep}{3pt}
			\setlength{\topsep}{3pt}
			\setlength{\partopsep}{0pt}
			\setlength{\leftmargin}{1.5em}
			\setlength{\labelwidth}{1em}
			\setlength{\labelsep}{0.5em} } }

	\newcounter{Lcount}
	\newcommand{\squishlisttwo}{
		\begin{list}{\arabic{Lcount}. }
			{ \usecounter{Lcount}
				\setlength{\itemsep}{0pt}
				\setlength{\parsep}{0pt}
				\setlength{\topsep}{0pt}
				\setlength{\partopsep}{0pt}
				\setlength{\leftmargin}{2em}
				\setlength{\labelwidth}{1.5em}
				\setlength{\labelsep}{0.5em} } }
		
		\newcommand{\squishend}{
	\end{list} }

\makeatletter
\def\adl@drawiv#1#2#3{%
	\hskip.5\tabcolsep
	\xleaders#3{#2.5\@tempdimb #1{1}#2.5\@tempdimb}%
	#2\z@ plus1fil minus1fil\relax
	\hskip.5\tabcolsep}
\newcommand{\cdashlinelr}[1]{%
	\noalign{\vskip\aboverulesep
		\global\let\@dashdrawstore\adl@draw
		\global\let\adl@draw\adl@drawiv}
	\cdashline{#1}
	\noalign{\global\let\adl@draw\@dashdrawstore
		\vskip\belowrulesep}}
\def\blfootnote{\xdef\@thefnmark{}\@footnotetext}
\makeatother

\aclfinalcopy % Uncomment this line for the final submission
\title{Dependency-Guided LSTM-CRF for Named Entity Recognition}

\author{Zhanming Jie \and Wei Lu\\
  StatNLP Research Group\\
  Singapore University of Technology and Design \\
  \texttt{zhanming\_jie@mymail.sutd.edu.sg, luwei@sutd.edu.sg} \\}
\date{}

\begin{document}
\maketitle
\begin{abstract}
%  Dependency tree structures leverage long-distance and syntactic relationships between the words in sentences.
  Dependency tree structures capture long-distance and syntactic relationships between words in a sentence.  
  The syntactic relations (e.g., {\it nominal subject}, {\it object}) can potentially infer the existence of certain named entities. 
  In addition, the performance of a named entity recognizer could benefit from the long-distance dependencies between the words in dependency trees. 
  In this work, we propose a simple yet effective {\it dependency-guided} LSTM-CRF model to encode the complete dependency trees and capture the above properties for the task of named entity recognition (NER).
%  present a thorough study of diverse neural architecture used for encoding the dependency tree and show the effectiveness of . 
%% describe the findings.
%  We conduct extensive experiments on several datasets to justify the key findings in this work. 
  The data statistics show strong correlations between the entity types and dependency relations.
  We conduct extensive experiments on several standard datasets and demonstrate the effectiveness of the proposed model in improving NER and achieving state-of-the-art performance. 
%  Our further experiments also show that the dependencies are helpful under a low-resource setting.
%  the performance of NER in low-resource language can be improved by the dependency-guided models. 
%  , especially in the scenario where we have limited amount of training data. 
  Our analysis reveals that the significant improvements mainly result from the dependency relations and long-distance interactions provided by dependency trees.
%  \footnote{We make our system and code available at *URL*.} 
%  The analysis discov ers frequent dependency patterns that contribute to the performance of NER. 
%  Our study suggests that high-quality dependency tree information is helpful in the NER tasks, especially for low-resource languages. \footnote{Code to reproduce the experiments is available at *URL*.}
  
\end{abstract}

\section{Introduction}
\label{sec:intro}
\blfootnote{Accepted as a long paper in EMNLP 2019 (Conference on Empirical Methods in Natural Language Processing).}
Named entity recognition (NER) is one of the most important and fundamental tasks in natural language processing (NLP).
Named entities capture useful semantic information which was shown helpful for downstream NLP tasks such as coreference resolution~\cite{lee2017end}, relation extraction~\cite{miwa2016end} and semantic parsing~\cite{dong2018coarse}.
%The named entity information can be applied to numerous downstream tasks, such as coreference resolution~\cite{lee2017end}, relation extraction~\cite{miwa2016end} and semantic parsing~\cite{dong2018coarse}.
%and sentiment analysis~\cite{mitchell2013open}. 
%Existing work~\cite{peters2018deep,devlin2019bert,akbik2018coling} has apply the deep contextualized word representation to achieve the state-of-the-art performance on standard NER corpora such as CoNLL-2003~\cite{tjong2003introduction} and OntoNotes~\cite{weischedel2013ontonotes}. 
%On the other hand, syntactic dependency trees which serve as discrete features~\cite{sasano2008japanese,cucchiarelli2001unsupervised,ling2012fine} or structural constraints~\cite{jie2017efficient} were shown to be effective on the NER task.
On the other hand, dependency trees also capture useful semantic information within natural language sentences. 
Currently, research efforts have derived useful discrete features from dependency structures~\cite{sasano2008japanese,cucchiarelli2001unsupervised,ling2012fine} or structural constraints~\cite{jie2017efficient} to help the NER task. 
%However, the dependency relations and  long-distance interactions which could benefit the NER performance are not completely captured by these approaches. 
However, how to make good use of the rich relational information as well as complex long-distance interactions among words as conveyed by the complete dependency structures for improved NER remains a research question to be answered.
%However, how to make good use of the rich relational information and complete dependency structures of sentences, which convey the long-distance interactions among words, 
%as well as complex long-distance interactions among words as conveyed by the complete dependency structures for improved NER remains a research question to be answered.
%Our observations show the benefits of such information leveraged from the dependencies. 
%In addition, syntactic dependency trees served as discrete features~\cite{sasano2008japanese,cucchiarelli2001unsupervised,ling2012fine} or structural constraints~\cite{jie2017efficient} were shown to be effective on this task. 

%11	they	_	PRP	PRP	_	12	nsubj	_	_	O
%12	suffer	_	VBP	VBP	_	10	ccomp	_	_	O
%13	from	_	IN	IN	_	12	prep	_	_	O
%14	drug	_	NN	NN	_	15	nn	_	_	O
%15	dealing	_	NN	NN	_	13	pobj	_	_	O
%16	and	_	CC	CC	_	15	cc	_	_	O
%17	loitering	_	NN	NN	_	15	conj	_	_	O
%18	near	_	IN	IN	_	15	prep	_	_	O
%19	their	_	PRP$	PRP$	_	20	poss	_	_	O
%20	premises	_	NNS	NNS	_	18	pobj	_	_	B-LOC
%21	.	_	.	.	_	10	punct	_	_	O
\begin{figure}[t!]
	\centering
	\adjustbox{max width=1.0\linewidth}{
	\begin{tikzpicture}[node distance=1.0mm and 1.0mm, >=Stealth, 
		wordnode/.style={draw=none, minimum height=5mm, inner sep=0pt},
		chainLine/.style={line width=1pt,-, color=fontgray},
		entbox/.style={draw=black, rounded corners, fill=red!20, dashed}
		]
%		\node [word](w1) [] {\footnotesize Ah};
%		\node [word, right=of w1](w2) [] {\footnotesize ,};
%		\node [word, right=of w2](w3) [] {\footnotesize today};
%		\node [word, right=of w3](w4) [] {\footnotesize is};
		\matrix (sent1) [matrix of nodes, nodes in empty cells, execute at empty cell=\node{\strut};]
		{
		 They & [1mm]suffer &[1mm]from & [1mm]drug   &  [1mm]  dealing & [1mm]and& [1mm]loitering& [1mm]near & [1mm]their & [1mm]premises\\
%		 \textbf{\textsc{date}}   &   &   & \textsc{o}     & \textsc{o}   & \textsc{o}  & \textsc{o}& \textsc{event} &  \\
		};
	
		\draw [chainLine, ->, color=fontgray, line width=1.5pt] (sent1-1-2) to [out=120,in=60, looseness=1.4] node[above, yshift=-1mm, color=black]{\footnotesize\it nsubj} (sent1-1-1);
		\draw [chainLine, ->] (sent1-1-2) to [out=60,in=120, looseness=1] node[above, yshift=-1mm, color=black]{\footnotesize\it prep} (sent1-1-3);
		\draw [chainLine, ->] (sent1-1-5) to [out=120,in=60, looseness=1] node[above, yshift=-1mm, xshift=-3mm, color=black]{\footnotesize\it prep} (sent1-1-4);
		\draw [chainLine, ->] (sent1-1-3) to [out=60,in=120, looseness=1.4] node[above, yshift=-1mm, color=black]{\footnotesize\it pobj} (sent1-1-5);
		\draw [chainLine, ->] (sent1-1-5) to [out=60,in=120, looseness=1] node[above, yshift=-1mm, color=black, xshift=2mm]{\footnotesize\it cc} (sent1-1-6);
		\draw [chainLine, ->] (sent1-1-5) to [out=60,in=120, looseness=1.5] node[above, yshift=-1mm, color=black, xshift=2mm]{\footnotesize\it conj} (sent1-1-7);
		\draw [chainLine, ->] (sent1-1-5) to [out=60,in=120, looseness=1.5] node[above, yshift=-1mm, color=black]{\footnotesize\it prep} (sent1-1-8);
		\draw [chainLine, ->, color=fontgray, line width=1pt] (sent1-1-10) to [out=120,in=60, looseness=1.4] node[above, yshift=-1mm, color=black, xshift=-3mm]{\footnotesize\it poss} (sent1-1-9);
		\draw [chainLine, ->, line width=1.5pt, color=myred] (sent1-1-8) to [out=60,in=120, looseness=1.5] node[above, yshift=-1mm, color=myred]{\footnotesize\textit{\textbf{pobj}} } (sent1-1-10);
		
		\begin{pgfonlayer}{background}
		\node [entbox, below=of sent1-1-10, yshift=7mm, text height=8mm, minimum width=15mm] (e1)  [] {\color{blue!80}\textbf{\textsc{loc}}};
%		\node [entbox, below=of sent1-1-8, xshift=5mm, yshift=6.1mm, text height=8mm, minimum width=18mm, fill=yellow!40] (e2)  [] {\color{blue!80}\textbf{\textsc{Event}}};
		\end{pgfonlayer}

		\matrix (sent2) [matrix of nodes, nodes in empty cells, execute at empty cell=\node{\strut};, below=of sent1, yshift=-14mm]
		{
			The &[-1mm] seminar &[-1mm] on &[-1mm] the & [-1mm]actual   &    [-1mm]practice &[-1mm] of & [-1mm]tax& [-1mm]reform & [-1mm]was & [-1mm]held & [-1mm]in &[-1mm] Hong &[-1mm] Kong\\
			%		 \textbf{\textsc{date}}   &   &   & \textsc{o}     & \textsc{o}   & \textsc{o}  & \textsc{o}& \textsc{event} &  \\
		};
	
		\draw [chainLine, ->] (sent2-1-2) to [out=120,in=60, looseness=1] node[above, yshift=-1mm, xshift=0mm,  color=black]{\footnotesize\it det} (sent2-1-1);
		\draw [chainLine, <-] (sent2-1-3) to [out=120,in=60, looseness=1.5] node[above, yshift=-1.5mm,  color=black, xshift=2mm]{\footnotesize\it prep} (sent2-1-2);
		\draw [chainLine, ->] (sent2-1-6) to [out=120,in=60, looseness=1.5] node[above, yshift=-1mm, xshift=-1mm,  color=black]{\footnotesize\it det} (sent2-1-4);
		\draw [chainLine, ->] (sent2-1-6) to [out=120,in=60, looseness=1] node[above, yshift=-1mm, xshift=-2mm,  color=black]{\footnotesize\it amod} (sent2-1-5);
		\draw [chainLine, ->] (sent2-1-3) to [out=60,in=120, looseness=1.6] node[above, yshift=-1.5mm,  color=black]{\footnotesize\it pobj} (sent2-1-6);
		\draw [chainLine, ->] (sent2-1-6) to [out=60,in=120, looseness=1.5] node[above, yshift=-1.5mm,  color=black]{\footnotesize\it prep} (sent2-1-7);
		\draw [chainLine,<-, line width=1pt] (sent2-1-8) to [out=60,in=120, looseness=1.5] node[above, yshift=-1mm, color=black,xshift=-1mm]{\footnotesize\it nn} (sent2-1-9);
		\draw [chainLine, ->] (sent2-1-7) to [out=60,in=120, looseness=1.8] node[above, yshift=-1.5mm,  color=black]{\footnotesize\it pobj} (sent2-1-9);
		
		\draw [chainLine,->, line width=1pt] (sent2-1-9) to [out=60,in=120, looseness=1.5] node[above, yshift=-1mm, color=black,xshift=-1mm]{\footnotesize\it auxpass} (sent2-1-10);
		\draw [chainLine, ->] (sent2-1-11) to [out=60,in=120, looseness=3.0] node[above, yshift=-1.5mm,  color=black]{\footnotesize\it prep} (sent2-1-12);
		
		\draw [chainLine, ->, color=myred, line width=1.5pt] (sent2-1-12) to [out=60,in=120, looseness=1.6] node[above, yshift=-1.5mm,  color=myred]{\footnotesize\textit{\textbf{pobj}}} (sent2-1-14);
		\draw [chainLine, <-] (sent2-1-13) to [out=60,in=120, looseness=1.4] node[above, yshift=-1mm,  color=black, xshift=-1mm]{\footnotesize\it nn} (sent2-1-14);
		
		\draw [chainLine, <-, color=myred, line width=1.5pt] (sent2-1-2) to [out=60,in=120, looseness=0.8] node[above, yshift=-1mm,  color=myred, xshift=-1mm]{\footnotesize\textit{\textbf{nsubjpass}}} (sent2-1-11);

		\begin{pgfonlayer}{background}
		\node [entbox, below=of sent2-1-5, xshift=5.7mm, yshift=5.8mm, text height=8mm, minimum width=72mm, fill=myblue!20] (e2)  [] {\color{blue!80}\textbf{\textsc{event}}};
		\node [entbox, below=of sent2-1-13, xshift=5mm, yshift=6.5mm, text height=8mm, minimum width=20mm, fill=yellow!40] (e2)  [] {\color{blue!80}\textbf{\textsc{gpe}}};
		\end{pgfonlayer}
	\end{tikzpicture} 
	}
%	\vspace*{-7.5mm}
	\caption{Example sentences annotated with named entiteis and dependencies in the OntoNotes 5.0 dataset.}
%	\vspace*{-6mm}
	\label{fig:examples}
\end{figure}
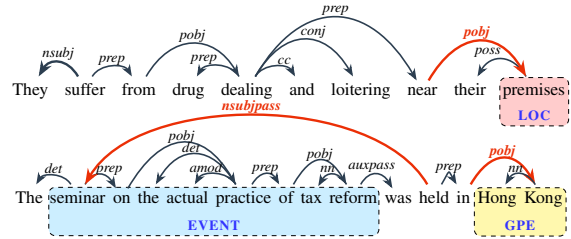

The first example in Figure \ref{fig:examples} illustrates the relationship between a dependency structure and a named entity. 
Specifically, the word ``{\it premises}'', which is a named entity of type \textsc{loc} (location), is characterized by the incoming arc with label ``{\it pobj}'' (prepositional object). 
This arc reveals a certain level of the semantic role that the word ``{\it premises}'' plays in the sentence. 
Similarly, the two words ``{\it Hong Kong}'' in the second example that form an entity of type \textsc{gpe} are also characterized by a similar dependency arc towards them.

The long-distance dependencies capturing non-local structural information can also be very helpful for the NER task~\cite{finkel2005incorporating}. 
In the second example of Figure \ref{fig:examples}, the long-distance dependency from ``\textit{held}'' to ``\textit{seminar}'' indicates a direct relation ``\textit{nsubjpass}'' (passive subject) between them, which can be used to characterize the existence of an entity. 
However, existing NER models based on linear-chain structures would have difficulties in capturing such long-distance relations (i.e., non-local structures).

One interesting property, as highlighted in the work of \citet{jie2017efficient}, is that most of the entities form subtrees under their corresponding dependency trees. 
In the example of the \textsc{Event} entity in Figure \ref{fig:examples}, the entity itself forms a subtree and the words inside have rich complex dependencies among themselves. 
Exploiting such dependency edges within the subtrees allows a model to capture non-trivial semantic-level interactions between words within long entities.
For example, ``\textit{practice}'' is the prepositional object (\textit{pobj}) of ``\textit{on}'' which is a preposition (\textit{prep}) of ``\textit{seminar}'' in the \textsc{Event} entity. 
Modeling these grandchild dependencies (GD)~\cite{koo2010efficient} requires the model to capture some higher-order long-distance interactions among different words in a sentence.

Inspired by the above characteristics of dependency structures, in this work, we propose a simple yet effective dependency-guided model for NER. 
Our neural network based model is able to capture both contextual information and rich long-distance interactions between words for the NER task. 
Through extensive experiments on several datasets on different languages, we demonstrate the effectiveness of our model, which achieves the state-of-the-art performance. 
To the best of our knowledge, this is the first work that leverages the complete dependency graphs for NER.
 We make our code publicly available at \url{http://www.statnlp.org/research/information-extraction}.

%0	Ah	UH	6	discourse	O	O
%1	,	,	6	punct	O	O
%2	today	NN	6	nsubj	S-DATE	S-DATE
%3	is	VBZ	6	cop	O	O
%4	the	DT	6	det	O	O
%5	first	JJ	6	amod	S-ORDINAL	S-ORDINAL
%6	workday	NN	-1	root	O	O
%7	after	IN	6	prep	O	O
%8	the	DT	11	det	O	O
%9	New	NNP	10	nn	B-EVENT	B-EVENT
%10	Year	NNP	11	nn	E-EVENT	E-EVENT
%11	holiday	NN	7	pobj	O	O
%12	.	.	6	punct	O	O

% long dependency
%8	Lee	NNP	10	nn	B-PERSON	B-PERSON
%9	Ann	NNP	10	nn	I-PERSON	I-PERSON
%10	Womack	NNP	7	pobj	E-PERSON	E-PERSON
%11	,	,	13	punct	O	O
%12	was	VBD	13	auxpass	O	O
%13	awarded	VBN	-1	root	O	O
%14	Single	NNP	13	dobj	B-WORK_OF_ART	B-WORK_OF_ART
%15	of	IN	14	prep	I-WORK_OF_ART	I-WORK_OF_ART
%16	the	DT	17	det	I-WORK_OF_ART	I-WORK_OF_ART
%17	Year	NNP	15	pobj	E-WORK_OF_ART	E-WORK_OF_ART
%18	and	CC	14	cc	O	O
%19	Song	NNP	14	conj	B-WORK_OF_ART	B-WORK_OF_ART
%20	of	IN	19	prep	I-WORK_OF_ART	I-WORK_OF_ART
%21	the	DT	22	det	I-WORK_OF_ART	I-WORK_OF_ART
%22	Year	NNP	20	pobj	E-WORK_OF_ART	E-WORK_OF_ART
%23	.	.	13	punct	O	O
% " I Hope you Dance " was awarded Song of the Year

\section{Related Work}
NER has been a long-standing task in the field of NLP. 
While many recent works~\cite{peters2018deep,akbik2018coling,devlin2019bert} focus on finding good contextualized word representations for improving NER, our work is mostly related to the literature that focuses on employing dependency trees for improving NER.

\citet{sasano2008japanese} exploited the syntactic dependency features for Japanese NER and achieved improved performance with a support vector machine~(SVM)~\cite{cortes1995support} classifier. 
Similarly, \citet{ling2012fine} included the head word in a dependency edge as features for fine-grained entity recognition. 
Their approach is a pipeline where they extract the entity mentions with  linear-chain conditional random fields~(CRF)~\cite{lafferty2001conditional} and used a classifier to predict the entity type.
%then classify the mentions into multiple entity labels. 
\citet{liu2010recognizing} proposed to link the words that are associated with selected typed dependencies (e.g., ``{\it nn}'', ``{\it prep}'') using a skip-chain CRF~\cite{sutton2004collective} model. 
They showed that some specific relations between the words can be exploited for improved NER. 
%As the underlying model involves loopy structures, exact inference is not available. 
%As the underlying model involves loopy structures, they performed inexact inference.
\citet{cucchiarelli2001unsupervised} applied a dependency parser to obtain the syntactic relations for the purpose of unsupervised NER. 
The resulting relation information serves as the features for potential existence of named entities. 
\citet{jie2017efficient} proposed an efficient dependency-guided model based on the semi-Markov CRF~\cite{sarawagi2004semi} for NER. 
The purpose is to reduce time complexity while maintaining the non-Markovian features.
%in the model. 
% properties of semi-Markov CRF. 
They observed certain relationships between the dependency edges and the named entities. 
Such relationships are able to define a reduced search space for their model. 
While these previous approaches do not make full use of the dependency tree structures, we focus on exploring neural architectures to exploit the  complete structural information conveyed by the dependency trees.
% dependency trees including the dependency relations. 

%Other than dependency trees, NER can also be benefited from other linguistics structures like parse trees, which are more expensive in terms of annotation costs compared to the dependency trees.  
%\citet{finkel2009joint} proposed a CRF-CFG model for joint parsing and named entity recognition. 
%Their model achieved competitive performance on the OntoNotes dataset~\cite{pradhan2013towards}. 
%\citet{li2017leveraging} applied the recursive neural network on top of the parse tree structures and obtained state-of-the-art performance on the OntoNotes dataset.

%We further analyze the useful dependency patterns for 

\section{Model}

Our dependency-guided model is based on the state-of-the-art BiLSTM-CRF model proposed by \citet{lample2016neural}.
% as it leverages the contextual representations which play an important role in the NER task. 
We first briefly present their model  as background and next present our dependency-guided model.
%and extended it to the dependency-guided models. 

\subsection{Background: BiLSTM-CRF}
In the task of named entity recognition, we aim to predict the label sequence $\vec{y} = \{y_1, y_2,\cdots, y_n\}$ given the input sentence $\vec{x}=\{x_1, x_2, \cdots, x_n\}$ where $n$ is the number of words.
%with $n$ words. 
The labels in $\vec{y}$ are defined by a label set with the standard \textsc{iobes}\footnote{``\textsc{s-}'' indicates the entity with a single word and ``\textsc{e-}'' indicates the end of an entity.} labeling scheme~\cite{ramshaw1999text,ratinov2009design}. 
The CRF~\cite{lafferty2001conditional} layer defines the probability of the label sequence $\vec{y}$ given $\vec{x}$:
\begin{equation}
P(\vec{y} \vert \vec{x})
=
\frac
{\exp ( score(\vec{x}, \vec{y})  )  }
{\sum_{\vec{y}^\prime }\exp ( score(\vec{x}, \vec{y}^\prime) )  }
\end{equation}
%\begin{center}
%	\begin{tabular}{c}
%		$
%		P(\vec{y} \vert \vec{x})
%		=
%		\frac
%		{\exp ( score(\vec{x}, \vec{y})  )  }
%		{\sum_{\vec{y}^\prime }\exp ( score(\vec{x}, \vec{y}^\prime) )  }
%		$
%	\end{tabular}
%	%\vspace{-2mm}
%\end{center}

Following \citet{lample2016neural}, the score is defined as the sum of transitions and emissions from the bidirectional LSTM (BiLSTM): 
\begin{equation}
score(\vec{x} , \vec{y}) = \sum_{i=0}^n A_{y_i, y_{i+1}} + \sum_{i=1}^n F_{\vec{x}, y_i}
\end{equation}
%\begin{center}
%	\begin{tabular}{c}
%		$
%		score(\vec{x} , \vec{y}) = \sum\limits_{i=0}^n A_{y_i, y_{i+1}} + \sum\limits_{i=1}^n F_{\vec{x}, y_i}
%		$
%	\end{tabular}
%	%\vspace{-2mm}
%\end{center}
where $\mathbf{A}$ is a transition matrix in which $A_{y_i, y_{i+1}}$ is the transition parameter from the label $y_i$ to the label $y_{i+1}$\footnote{$y_0$ and $y_{n+1}$ are \texttt{start} and \texttt{end} labels.}. 
$\mathbf{F}_\vec{x}$ is an emission matrix where $F_{\vec{x}, y_i}$ represents the scores of the label  $y_i$ at the $i$-th position. 
Such scores are provided by the parameterized LSTM~\cite{hochreiter1997long} networks. 
During training, we minimize the negative log-likelihood to obtain the model parameters including both LSTM and transition parameters. 
%During training, we minimize the The model parameters can be obtained by minimizing the negative log-likelihood.

\subsection{Dependency-Guided LSTM-CRF}
\paragraph{Input Representations}
The word representation $\vec{w}$ in the BiLSTM-CRF~\cite{lample2016neural,ma2016end,D17-1035} model consists of the concatenation of the word embedding as well as the corresponding character-based representation. 
Inspired by the fact that each word (except the root) in a sentence has exactly one \textit{head} (i.e., \textit{parent}) word in the dependency structure, we can enhance the word representations with such dependency information. 
%Specifically, the head word embeddings and dependency relation embeddings can be included as part to the word representation. 
Similar to the work by \citet{miwa2016end}, we concatenate the word representation together with the corresponding head word representation and dependency relation embedding as the input representation. 
Specifically, given a dependency edge $(x_h, x_i, r)$ with $x_h$ as parent, $x_i$ as child and $r$ as dependency relation, the representation at position $i$ can be denoted as:
\begin{equation}
\vec{u}_i = \left [
\vec{w}_i; \vec{w}_h; \vec{v}_{r}	
\right ], ~~ x_h = parent(x_i)
\label{equ:rep}
\end{equation}
%\begin{center}
%	\begin{tabular}{c}
%		$
%		\vec{u}_i = \left [
%		\vec{w}_i; \vec{w}_h; \vec{v}_{r}	
%		\right ], ~~ x_h = parent(x_i)
%		$
%	\end{tabular}
%	%\vspace{-2mm}
%\end{center}
where $\vec{w}_i$ and $\vec{w}_h$ are the word representations of the word $x_i$ and its parent $x_h$, respectively. 
%$\vec{w}_i$ is the concatenation of pre-trained word embeddings (e.g. Glove~\cite{pennington2014glove}) of $x_i$ and its character-based representation. 
We take the final hidden state of character-level BiLSTM as the character-based representation~\cite{lample2016neural}.
%Following state-of-the-art neural architectures~\cite{lample2016neural,ma2016end,D17-1035}, a word representation is the concatenation of pre-trained word embeddings (e.g. Glove~\cite{pennington2014glove}) and its character-based word presentation $\vec{c}_i$. 
$\vec{v}_r$ is the embedding for the dependency relation $r$. 
These relation embeddings are randomly initialized and fine-tuned during training. 
The above representation allows us to capture the direct long-distance interactions at the input layer. 
For the word that is a root of the dependency tree, we treat its parent as itself\footnote{We also tried using a root word embedding but the performance is similar.} and create a root relation embedding. 
Additionally, contextualized word representations (e.g., ELMo) can also be concatenated into $\vec{u}$.

\begin{figure}[t!]
	\centering
	\adjustbox{max width=1\linewidth}{
		\begin{tikzpicture}[node distance=8.0mm and 10mm, >=Stealth, 
		sentity/.style={draw=none, circle, minimum height=8.5mm, minimum width=8.5mm,line width=1pt, inner sep=0pt, fill=purple!35},
		lstm/.style={draw=none, minimum height=5mm, rounded rectangle, fill=puffin!50, minimum width=11.5mm, label={center:\tiny \textcolor{fontgray}{\bf LSTM}}},
		%		, blur shadow={shadow blur steps=5}
		plus/.style={draw=none, minimum height=3mm, circle, fill={rgb:red,0;green,210;blue,211}, label={center:\footnotesize \textcolor{white}{+}}},
		gfunc/.style={draw=none, minimum height=5mm, rectangle, fill=green!20, label={center:\footnotesize \textcolor{fontgray}{$g(\cdot)$}}, minimum width=8mm, line width = 1.5pt},
		emb/.style={draw=none, minimum height=3mm, rounded rectangle, fill=none, minimum width=12mm, text=fontgray},
		%		cyan!40
		chainLine/.style={line width=0.8pt,->, color=mygray},
		%	background rectangle/.style={fill=olive!45}, show background rectangle, 
		%		line width=1.3pt
		]

		\matrix (s) [matrix of nodes, nodes in empty cells, execute at empty cell=\node{\strut};]
		{
			Abramov & [2ex]had & [5ex]an & [3ex]accident   &   [3ex] in  & [3ex] Moscow\\
			%		 \textbf{\textsc{date}}   &   &   & \textsc{o}     & \textsc{o}   & \textsc{o}  & \textsc{o}& \textsc{event} &  \\
		};
		
		\draw [chainLine, ->] (s-1-2) to [out=120,in=60, looseness=1] node[above, yshift=-1mm, color=black]{\footnotesize\it nsubj} (s-1-1);
		\draw [chainLine, ->] (s-1-4) to [out=120,in=60, looseness=0.8] node[above, yshift=-1mm, color=black]{\footnotesize\it det} (s-1-3);
		\draw [chainLine, ->] (s-1-2) to [out=60,in=120, looseness=0.9] node[above, yshift=-1.2mm, xshift=3mm, color=black]{\footnotesize\it dobj} (s-1-4);
		\draw [chainLine, ->] (s-1-2) to [out=60,in=120, looseness=0.9] node[above, yshift=-1mm, color=black]{\footnotesize\it prep} (s-1-5);
		\draw [chainLine, ->] (s-1-5) to [out=60,in=120, looseness=1] node[above, yshift=-1mm, color=black]{\footnotesize\it pobj} (s-1-6);
		
		\node[emb, above= of s-1-1, yshift=5mm] (u1) {$\vec{u}_\mathbf{1}$};
		\node[emb, above= of s-1-2, yshift=5mm] (u2) {$\vec{u}_\mathbf{2}$};
		\node[emb, above= of s-1-3, yshift=5.8mm] (u3) {$\vec{u}_\mathbf{3}$};
		\node[emb, above= of s-1-4, yshift=5mm] (u4) {$\vec{u}_\mathbf{4}$};
		\node[emb, above= of s-1-5, yshift=5mm] (u5) {$\vec{u}_\mathbf{5}$};
		\node[emb, above= of s-1-6, yshift=5mm] (u6) {$\vec{u}_\mathbf{6}$};
		
		\node[lstm, above= of u1] (h1) {};
		\node[lstm, above=of u2](h2) {};
		\node[lstm, above= of u3](h3) {};
		\node[lstm, above= of u4](h4) {};
		\node[lstm, above= of u5](h5) {};
		\node[lstm, above= of u6](h6) {};

		\node[gfunc, above= of h1, yshift=-1mm] (g1) {};
		\node[gfunc, above=of h2, yshift=-1mm](g2) {};
		\node[gfunc, above= of h3, yshift=-1mm](g3) {};
		\node[gfunc, above= of h4, yshift=-1mm](g4) {};
		\node[gfunc, above= of h5, yshift=-1mm](g5) {};
		\node[gfunc, above= of h6, yshift=-1mm](g6) {};
		
		\node[lstm, above= of g1, yshift=-3mm] (m1) {};
		\node[lstm, above=of g2, yshift=-3mm](m2) {};
		\node[lstm, above= of g3, yshift=-3mm](m3) {};
		\node[lstm, above= of g4, yshift=-3mm](m4) {};
		\node[lstm, above= of g5, yshift=-3mm](m5) {};
		\node[lstm, above= of g6, yshift=-3mm](m6) {};
		
		%		\node[left = of m1, xshift](layer1) {$L=1$};
		
		\node[sentity, above= of m1] (e1) {\footnotesize \textcolor{fontgray}{\bf \textsc{s-per}}};
		\node[sentity, above=of m2](e2) {\footnotesize \textcolor{fontgray}{\bf \textsc{o}}};
		\node[sentity, above= of m3](e3) {\footnotesize \textcolor{fontgray}{\bf \textsc{o}}};
		\node[sentity, above= of m4](e4) {\footnotesize \textcolor{fontgray}{\bf \textsc{o}}};
		\node[sentity, above= of m5](e5) {\footnotesize \textcolor{fontgray}{\bf \textsc{o}}};
		\node[sentity, above= of m6](e6) {\footnotesize \textcolor{fontgray}{\bf \textsc{s-gpe}}};

		\draw [chainLine] (u1) to (h1);
		\draw [chainLine] (u2) to (h2);
		\draw [chainLine] (u3) to (h3);
		\draw [chainLine] (u4) to (h4);
		\draw [chainLine] (u5) to (h5);
		\draw [chainLine] (u6) to (h6);
		
		\draw [chainLine] (h1) to  (g1);
		\draw [chainLine] (h2) to (g2);
		\draw [chainLine] (h3) to (g3);
		\draw [chainLine] (h4) to (g4);
		\draw [chainLine] (h5) to (g5);
		\draw [chainLine] (h6) to (g6);
		
		\draw [chainLine] (g1) to (m1);
		\draw [chainLine] (g2) to (m2);
		\draw [chainLine] (g3) to (m3);
		\draw [chainLine] (g4) to (m4);
		\draw [chainLine] (g5) to (m5);
		\draw [chainLine] (g6) to (m6);
		
		\draw [chainLine] (m1) to  (e1);
		\draw [chainLine] (m2) to   (e2);
		\draw [chainLine] (m3) to   (e3);
		\draw [chainLine] (m4) to    (e4);
		\draw [chainLine] (m5) to   (e5);
		\draw [chainLine] (m6) to   (e6);
		
		\draw [chainLine, dashed] (h2) to  [out=60,in=-60] (g2);
		\draw [chainLine, dashed] (h2) to (g1);
		\draw [chainLine, dashed] (h2) to (g4);
		\draw [chainLine, dashed] (h2) to (g5);
		\draw [chainLine, dashed] (h4) to (g3);
		\draw [chainLine, dashed] (h5) to (g6);
		
		\draw [chainLine, <->] (h1) to (h2);
		\draw [chainLine, <->] (h2) to (h3);
		\draw [chainLine, <->] (h3) to (h4);
		\draw [chainLine, <->] (h4) to (h5);
		\draw [chainLine, <->] (h5) to (h6);
		
		\draw [chainLine, <->] (m1) to (m2);
		\draw [chainLine, <->] (m2) to (m3);
		\draw [chainLine, <->] (m3) to (m4);
		\draw [chainLine, <->] (m4) to (m5);
		\draw [chainLine, <->] (m5) to (m6);
		
		\draw [chainLine, line width=1.2pt, -, middlefactor] (e1) to (e2);
		\draw [chainLine, line width=1.2pt, -, middlefactor] (e2)to (e3);
		\draw [chainLine, line width=1.2pt, -, middlefactor] (e3)to (e4);
		\draw [chainLine, line width=1.2pt, -, middlefactor]  (e4)to (e5);
		\draw [chainLine, line width=1.2pt, -, middlefactor] (e5)to (e6);
		
		\end{tikzpicture} 
	}
%	\vspace*{-9mm}
	\caption{Dependency-guided LSTM-CRF with 2 LSTM Layers. Dashed connections mimic the dependency edges. ``$g(\cdot)$'' represents the interaction function.}
%	\vspace*{-5mm}
	\label{fig:model2layer}
\end{figure}
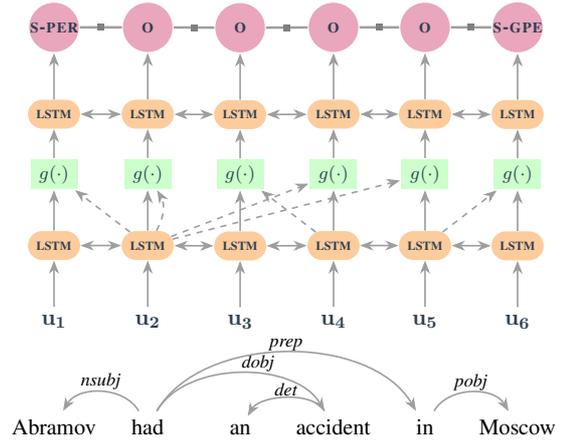

%1	Abramov	_	NNP	NNP	_	2	nsubj	_	_	B-PERSON
%2	had	_	VBD	VBD	_	0	root	_	_	O
%3	a	_	DT	DT	_	5	det	_	_	O
%4	car	_	NN	NN	_	5	nn	_	_	O
%5	accident	_	NN	NN	_	2	dobj	_	_	O
%6	in	_	IN	IN	_	2	prep	_	_	O
%7	Moscow	_	NNP	NNP	_	6	pobj	_	_	B-GPE
%8	last	_	JJ	JJ	_	9	amod	_	_	B-TIME
%9	night	_	NN	NN	_	2	tmod	_	_	I-TIME
%10	and	_	CC	CC	_	2	cc	_	_	O
%11	was	_	VBD	VBD	_	13	auxpass	_	_	O
%12	seriously	_	RB	RB	_	13	advmod	_	_	O
%13	injured	_	VBN	VBN	_	2	conj	_	_	O
%14	.	_	.	.	_	2	punct	_	_	O

\paragraph{Neural Architecture}
Given the dependency-encoded input representation $\vec{u}$, we apply the LSTM to capture the contextual information and model the interactions between the words and their corresponding parents in the dependency trees. 
%use dependencies to provide long-distance interactions guidance. 
Figure \ref{fig:model2layer} shows the proposed dependency-guided LSTM-CRF (\textbf{DGLSTM-CRF}) with 2 LSTM layers for the example sentence ``\textit{Abramov had an accident in Moscow}'' and its dependency structure. 
%Figure \ref{fig:model2layer} shows the overall dependency-guided LSTM-CRF ($\textbf{DGLSTM-CRF}$) model architecture given the example sentence ``\textit{Abramov had an accident in Moscow}'' and its dependency structure. 
The corresponding label sequence is \{\textsc{s-per}, \textsc{o}, \textsc{o}, \textsc{o}, \textsc{o}, \textsc{s-gpe}\}. 
%Followed by the first BiLSTM, the hidden states at each position will propagate to the next BiLSTM layer and propagate to its child following the dependency trees. 
Followed by the first BiLSTM, the hidden states at each position will propagate to the next BiLSTM layer and its child along the dependency trees. 
For example, the hidden state of the word ``\textit{had}'', $\vec{h}_2^{(1)}$, will propagate to its child ``\textit{Abramov}'' at the first position. 
For the word that is root, the hidden state at that specific position will propagate to itself. 
We use an \textit{interaction function} $g(\vec{h}_i, \vec{h}_{p_i})$ to capture the interaction between the child and its parent in a dependency. 
Such an interaction function can be concatenation, addition or a  multilayer perceptron (MLP). 
We further apply another BiLSTM layer on top of the interaction functions to produce the context representation for the final CRF layer. 

The architecture shown in Figure \ref{fig:model2layer} with a 2-layer BiLSTM can effectively encode the grandchild dependencies because the input representations encode the parent information and the interaction function further propagate the grandparent information. 
Such propagations allow the model to capture the indirect long-distance interactions from the grandchild dependencies between the words in the sentence as mentioned in Section \ref{sec:intro}. 
In general, we can stack more interaction functions and BiLSTMs to enable deeper reasoning over the dependency trees. 
%Following the dependency structure, we can obtain the representation $\vec{u}$ at each position (Equation \ref{equ:rep}).
%However, 1-layer LSTM may not be able to capture the grandchild dependencies, which are statistically related to named entity as mentioned Section \ref{sec:intro}. 
%The grandchild dependencies can capture longer interactions between the words in the sentence.
%The grandparent of an entity component could be part of the entity or has a direct relation to this entity. 
%To enable such interactions, we stack multiple layers of LSTMs and guide the flow of the representations to there parents at next layer. 
Specifically, the hidden states of the $(l+1)$-th layer $\mathbf{H}^{(l+1)}$ can be calculated from the hidden state of the previous layer $\mathbf{H}^{(l)}$:
\begin{equation}
\begin{split}
\mathbf{H}^{(l+1)} & \!   = \!\text{BiLSTM}
\Big ( 
f \big (\mathbf{H}^{(l)} \big )
\Big ) \\
\mathbf{H}^{(l)} & \! = \!\!
\left [
\vec{h}_1^{(l)}, \vec{h}_2^{(l)}, \cdots, \vec{h}_n^{(l)}
\right ] \\
f\big (\mathbf{H}^{(l)} \big ) &\!  = \!\!
\Big [
g (
\vec{h}_1^{(l)}, \vec{h}_{p_1}^{(l)}
), 
\cdots ,
g (
\vec{h}_n^{(l)},\vec{h}_{p_n}^{(l)}
)
\Big ]
\end{split}\nonumber
\end{equation}
%\begin{equation}
%\begin{split}
%\mathbf{H}^{(l+1)} & \!   = \!\text{BiLSTM}
%\Big ( 
%f \big (\mathbf{H}^{(l)} \big )
%\Big ) \\
%\mathbf{H}^{(l)} & \! = \!\!
%\left [
%\vec{h}_1^{(l)}, \vec{h}_2^{(l)}, \cdots, \vec{h}_n^{(l)}
%\right ] \\
%f\big (\mathbf{H}^{(l)} \big ) &\!  = \!\!
%\Big [
%   g (
%      \vec{h}_1^{(l)}, \vec{h}_{p_1}^{(l)}
%   ), 
%   \cdots ,
%  g (
%   \vec{h}_n^{(l)},\vec{h}_{p_n}^{(l)}
%   )
%\Big ]
%\end{split}\nonumber
%\end{equation}
where $p_i$ indicates the parent index of the word $x_i$. 
$g(\vec{h}_i^{(l)}, \vec{h}_{p_i}^{(l)})$ represents the interaction functions between the hidden state at the $i$-th and  $p_i$-th positions under the dependency edges $(x_{p_i}, x_i)$. 
%The interaction function is similar to the concept in 
%At the output of the first-layer LSTM in Figure \ref{fig:model2layer}, we concatenate the hidden state at position $i$ and the hidden state of its parent to make a new representation, which will be the input representation to the next layer. 
The number of layers $L$ can be chosen according to the performance on the development set. 

\begin{table}[t!]
	\centering
	\scalebox{0.8}{
		\begin{tabular}{ll}
			\toprule
			\textbf{Interaction Function}& $g(\vec{h}_i, \vec{h}_{p_i}) $ \\
			\midrule
			Self connection & $\vec{h}_i$\\
			Concatenation & $\vec{h}_i \bigoplus  \vec{h}_{p_i}$\\
			Addition & $\vec{h}_i +  \vec{h}_{p_i}$\\
			MLP & $\text{ReLU} \big (\mathbf{W}_1\vec{h}_i \! + \! \mathbf{W}_2\vec{h}_{p_i} \big )$ \\
			\bottomrule
		\end{tabular}
	}
%	\vspace*{-3mm}
	\caption{List of interaction functions.}
%	\vspace*{-4mm}
	\label{tab:interfunc}
\end{table}

\paragraph{Interaction Function}
The interaction function between the parent and child representations can be defined in various ways. 
Table \ref{tab:interfunc} shows the list of interaction function considered in our experiments. 
The first one returns the hidden state itself, which is equivalent to stacking the LSTM layers. 
%We investigate the following functions in the underlying model:
%\begin{equation}
%g(\vec{h}_i, \vec{h}_{p_i}) = 
%\left\{
%\begin{matrix}
%\vec{h}_i &  \! \!\!\!  \text{Self}\\ 
%\vec{h}_c \bigoplus  \vec{h}_p &  \! \!\!\!  \text{Concatenation}\\ 
%\vec{h}_c +  \vec{h}_p & \! \!\!\!  \text{Addition}\\ 
%\sigma \big (\mathbf{W}_1\vec{h}_c \! + \! \mathbf{W}_2\vec{h}_p \big ) & \!\! \!\!  \text{MLP} \\
%\end{matrix}\right.
%\nonumber
%\end{equation}
The concatenation and addition involve no parameter, which are straightforward ways to model the interactions. 
The last one applies an MLP to model the interaction between parent and child representations. 
%The interaction function $g(\vec{h}_c, \vec{h}_p)$ can take one of these 3 operations. 
%Specifically, the first two operations are intuitive and no parameter involve. 
%The last one is applying an MLP to model the interactions between parent and child representations. 
%$\sigma$ is the non-linear activation function. 
With the rectified linear unit (ReLU) as activation function, the $g(\vec{h}_i, \vec{h}_{p_i})$ function is analogous to a graph convolutional network (GCN)~\cite{kipf2017semi} formulation. 
In such a graph, each node has a self connection (i.e., $\vec{h}_i$) and a dependency connection with parent (i.e., $\vec{h}_{p_i}$).  
Similar to the work by \citet{marcheggiani2017encoding}, we adopt different parameters $\mathbf{W}_1$ and $\mathbf{W}_2$ for self and dependency connections. 

\section{Experiments}

\paragraph{Datasets}
The main experiments are conducted on the large-scale OntoNotes 5.0~\cite{weischedel2013ontonotes} English and Chinese datasets.
We chose these datasets  because  they contain both constituency tree and named entity annotations. 
There are 18 types of entities defined in the OntoNotes dataset. 
We convert the constituency trees into the Stanford dependency~\cite{de2008stanford} trees using the rule-based tool~\cite{de2006generating} by Stanford CoreNLP~\cite{manning2014stanford}.
For English, \citet{pradhan2013towards} provided the train/dev/test split\footnote{http://cemantix.org/data/ontonotes.html} and the split has been used by several previous works~\cite{chiu2016named,li2017leveraging,ghaddar2018robust}.
%for the CoNLL-2012 shared task~\cite{pradhan2012conll} and the split has been used by several previous works~\cite{chiu2016named,li2017leveraging,ghaddar2018robust}.
For Chinese, we use the official splits provided by \citet{pradhan2012conll}\footnote{http://conll.cemantix.org/2012/data.html}.
%For Chinese, we use the official splits provided by the CoNLL-2012 shared task\footnote{http://conll.cemantix.org/2012/data.html}, which is same as the one reported in \citet{pradhan2012conll}.
%The OntoNotes corpus contain texts from a wide range of sources, such as broadcast conversation, broadcast news, newswire, magazine, telephone converation, and Web text. 

\begin{table}[t!]
	\centering
	\setlength{\tabcolsep}{2pt} % Default value: 6pt
	\renewcommand{\arraystretch}{1.1} % Default value: 1
	\resizebox{1.0\linewidth}{!}{
		\begin{tabular}{l cccccccc}
			\toprule
			\multirow{2}{*}{\textbf{Dataset}} & \multicolumn{2}{c}{\bf Train} & \multicolumn{2}{c}{\bf Dev} & \multicolumn{2}{c}{\bf Test} & \textbf{ST} & \textbf{GD} \\[-1mm]
			\cmidrule(lr){2-3} \cmidrule(lr){4-5} \cmidrule(lr){6-7}
			& \textbf{\# Sent.} & \textbf{\# Entity} & \textbf{\# Sent.} & \textbf{\# Entity} & \textbf{\# Sent.} & \textbf{\# Entity} & \textbf{(\%)} & \textbf{(\%)} \\
			\midrule
			OntoNotes 5.0 - English & 59,924 & 81,828 & 8,528 & 11,066 & 8,262 & 11,057 & \textcolor{white}{0}98.5 & 41.1  \\
			OntoNotes 5.0 - Chinese & 36,487 & 62,543 & 6,083 & \textcolor{white}{0}9,104 & 4,472 & \textcolor{white}{0}7,494 & \textcolor{white}{0}92.9 & 49.1\\
			%			CoNLL-2003 English & 14,041 & 23,499 & 3,250 & 5,942 & 3,453& 5,648 & \textcolor{white}{0}92.1 & 34.4 \\ 
%			SemEval2010T1 - English & \textcolor{white}{0}3,648 & \textcolor{white}{0}7,625 & \textcolor{white}{0,}741 & \textcolor{white}{0}1,600  & 1,141 &  \textcolor{white}{0}2,474 & \textcolor{white}{0}89.3 & 26.5\\
			SemEval2010T1 - Catalan & \textcolor{white}{0}8,709 & 15,278 & 1,445 & \textcolor{white}{0}2,431 & 1,698 & \textcolor{white}{0}2,910 & 100.0 & 28.6 \\
			SemEval2010T1 - Spanish  & \textcolor{white}{0}9,022 & 17,297 & 1,419 & \textcolor{white}{0}2,615 & 1,705 & \textcolor{white}{0}3,046 & 100.0 & 29.8\\
			%		 \midrule \midrule 
			%		 LowResource - Afrikaans&  &  &  &  &  &  \\
			%		 LowResource - Indonesian &  &  &  &  &  &  \\
			%		 LowResource - Swedish &  &  &  &  &  &  \\
			\bottomrule
		\end{tabular}
	}
%	\vspace*{-3mm}
	\caption{Dataset statistics. ``ST'' is the ratio of entities that form subtrees. ``GD'' is the ratio of entities that have grandchild dependencies within their subtrees.}
%	\vspace*{-4mm}
	\label{tab:datastat}
\end{table}

Besides, we also conduct experiments on the Catalan and Spanish datasets from the SemEval-2010 Task 1\footnote{http://stel.ub.edu/semeval2010-coref/download}~\cite{recasens2010semeval}\footnote{This dataset also has English portion but it is a subset of the OntoNotes English.}. 
The SemEval-2010 task was originally designed for the task of coreference resolution in multiple languages. 
Again, we chose these corpora primarily because they contain both dependency and named entity annotations.
%These corpora contain both dependency and named entity annotations. 
Following \citet{finkel2009joint} and \citet{jie2017efficient}, we select the most dominant three entity types and merge the rest into one general a entity type ``\textit{misc}''. 
%less dominant entity types into ``\textit{misc}'' and categorize all entities into four types. 
Table \ref{tab:datastat} shows the statistics of the datasets used in main experiments. 
To further evaluate the effectiveness of the dependency structures, we also conduct additional experiments under a low-resource setting for NER~\cite{cotterell2017low}. 

The last two columns of Table \ref{tab:datastat} show the relationships between the dependency trees and named entities with length larger than 2 for the complete dataset. 
Specifically, the penultimate column shows the percentage of entities that can form a complete subtree (ST) under their dependency tree structures. 
Apparently, most of the entities form subtrees, especially for the Catalan and Spanish datasets where 100\% entities form subtrees.  
This observation is consistent with the findings reported in \citet{jie2017efficient}. 
%As introduced earlier in Section \ref{sec:intro}, we calculate the percentage of above entities that have the grandchild dependencies~\cite{koo2010efficient} (GD).
%As introduced earlier in Section \ref{sec:intro}, most of the entities themselves form dependency subtrees. 
The last column in Table \ref{tab:datastat} shows the percentage of the grandchild dependencies~\cite{koo2010efficient} (GD) that exist in these subtrees (i.e., entities).  
%The number showing in table is calculated under the constraint that both parent and grandparent in a grandchild dependency should be part of the entity. 
Such grandchild dependencies could be useful for detecting certain named entities, especially for long entities.
%capturing indirect long-distance interactions among the words, which could be u
%especially for long entities. 
As we will see later in Section \ref{sec:analysis}, the performance of long entities can be significantly improved with our dependency-guide model.

\begin{figure}[t!]
	\centering
	\scalebox{1.0}{
	\includegraphics[width=3.2in]{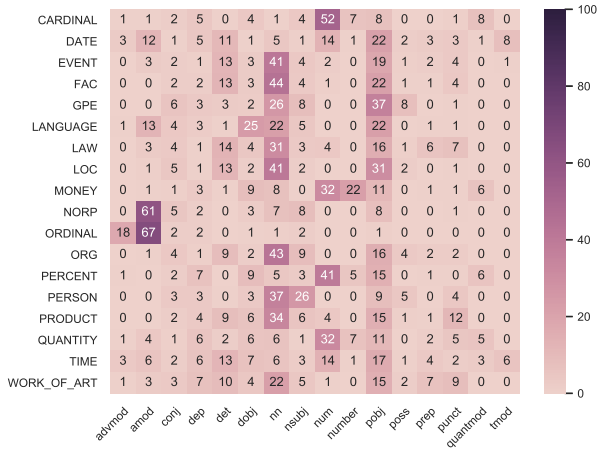}
}
	\vspace{-8mm}
	\caption{Percentage of entity words ($y$ axis) with respect to dependency relations ($x$ axis) in the OntoNotes English dataset. Columns with percentage less than 5\% are ignored for brevity.}
%	\vspace*{-5mm}
	\label{fig:depstat}
\end{figure}

%The heatmap table in Figure \ref{fig:depstat} shows the percentage of entity words\footnote{The words that are annotated with entity labels.} with respect to dependency relations in the OntoNotes English dataset. 
The heatmap table in Figure \ref{fig:depstat} shows the correlation between the entity types and the dependency relations in the OntoNotes English dataset. 
Specifically, each entry denotes the percentage of the entities that have a parent dependency with a specific dependency relation.  
For example, at the row with \textsc{gpe} entity, 37\% of the entity words\footnote{The words that are annotated with entity labels.} have a dependency edge whose label is ``\textit{pobj}''. 
When looking at column of ``\textit{pobj}'' and ``\textit{nn}'', we can see that most of the entities relate to the prepositional object (\textit{pobj}) and noun compound modifier (\textit{nn}) dependencies. 
Especially for the \textsc{norp} (i.e., {\it nationalities or religious or political groups}) and \textsc{ordinal} (e.g., ``\textit{first}'', ``\textit{second}'') entities, more than 60\% of the entity words have the dependency with  adjectival modifier (\textit{amod}) relation. 
%are related to the adjectival modifier (\textit{amod}) dependency. 
Furthermore, every entity type (i.e., row) has a most related dependency relation (with more than 17\% occurrences). 
Such observations present useful information that can be used to categorize named entities with different types. 
%Such observations show that dependency relations are good indicators of named entities. 

\paragraph{Baselines}
We implemented the state-of-the-art NER model \textbf{BiLSTM-CRF}~\cite{lample2016neural} as the first baseline with different number of LSTM layers ($L=\{0, 1,2,3\}$). 
$L=0$ indicates the model only relies on the input representation. 
%at most two LSTM layers ($L\leq2$). 
Following \citet{zhang2018graph}, the complete dependency trees are considered bidirectional and encoded with a contextualized GCN (BiLSTM-GCN). 
We further add the relation-specific parameters~\cite{marcheggiani2017encoding} and a CRF layer for the NER task. 
% we encode a complete dependency tree with GCN and add a CRF layer for the NER task. 
The resulting baseline is \textbf{BiLSTM-GCN-CRF}
\footnote{Detailed description of this baseline can also be found in the supplementary material.}. 
We use the bootstrapping paired t-test~\cite{berg2012empirical} for significance test when comparing the results of different models.
%Details regarding the baseline models can be found in the supplementary material. 

\paragraph{Experimental Setup}
%We implemented the state-of-the-art NER model BiLSTM-CRF as baseline~\cite{lample2016neural} with at most two LSTM layers. 
%The number of LSTM layers $L$ in DGLSTM-CRF is set to 2 and the number of GCN layers $J$ in DGGCN-CRF is set to 1. 
We choose MLP as the interaction function in our DGLSTM-CRF according to performance on the development set. 
The hidden size of all models (i.e., LSTM, GCN) is set to 200. 
We use the Glove~\cite{pennington2014glove} 100-d word embeddings, which was shown to be effective in English NER task~\cite{ma2016end,peters2018deep}. 
We use the publicly available FastText~\cite{grave2018learning} word embeddings for Chinese, Catalan and Spanish. 
%For Chinese, Catalan and Spanish, we use the FastText~\cite{grave2018learning} embeddings which are trained on Common Crawl and Wikipedia using CBOW. 
%as they provide \textcolor{red}{here}. 
%for English and FastText~\cite{grave2018learning} embeddings for the other languages.
%We use Glove~\cite{pennington2014glove} word embeddings for English and FastText~\cite{grave2018learning} embeddings for the other languages.
The ELMo~\cite{peters2018deep}, deep contextualized word representations\footnote{We also tried BERT~\cite{devlin2019bert} in preliminary experiments and obtained similar performance as ELMo. The NER performance using BERT without fine-tuning reported in \citet{peters2019tune} is consistent with the one reported by ELMo~\cite{peters2018deep}.} are used for all languages in our experiments since \citet{che2018towards} provides ELMo for many other languages\footnote{https://github.com/HIT-SCIR/ELMoForManyLangs}, including Chinese, Catalan and Spanish. 
We use the average weights over all layers of the ELMo representations and concatenate them with the input representation $\vec{u}$. 
Our models are optimized by mini-batch stochastic gradient descent (SGD) with learning rate 0.01 and batch size 10.
The $L_2$ regularization parameter is 1e-8. 
The hyperparameters are selected according to the performance on the OntoNotes English development set.

\subsection{Main Results}

\paragraph{OntoNotes English}
Table \ref{tab:mainresults} shows the performance comparison between our work and previous work on the OntoNotes English dataset. 
Without the LSTM layers (i.e., $L=0$), the proposed model with dependency information significantly improves the NER performance with more than 2 points in F$_1$ compared to the baseline BiLSTM-CRF ($L=0$), which demonstrate the effectiveness of dependencies for the NER task.
Our best performing BiLSTM-CRF baseline (with Glove) achieves a F$_1$ score of 87.78 which is better than or on par with previous works~\cite{chiu2016named,li2017leveraging,ghaddar2018robust} with extra features. 
This baseline also outperforms the CNN-based models~\cite{strubell2017fast,li2017leveraging}. 
The BiLSTM-GCN-CRF model outperforms the BiLSTM-CRF model but achieves inferior performance compared to the proposed DGLSTM-CRF model. 
We believe it is challenging to preserve the surrounding context information with stacking GCN layers while contextual information is important for NER~\cite{peters2018dissecting}.
%Because stacking GCN layers is challenging to preserve the surrounding context information, which is important for the task of NER~\cite{peters2018dissecting}.
%Although BiLSTM-GCN-CRF also encoded the complete dependency trees, the last GCN layer is lack of the ability to capture the surrounding context information, which is important for the task of NER~\cite{peters2018dissecting}. 
Overall, the 2-layer DGLSTM-CRF model significantly (with $p < 0.01$) outperforms the best BiLSTM-CRF baseline and the BiLSTM-GCN-CRF model. 
As we can see from the table, increasing the number of layers (e.g., $L$ = $3$) does not give us further improvements for both BiLSTM-CRF and DGLSTM-CRF because such third-order information (e.g., the relationship among a word’s parent, its grandparent, and great-grandparent) does not play an important role in indicating the presence of named entities.
% the performance for all the models. 
%The inferior performance of BiLSTM-GCN-CRF compared to DGLSTM-CRF attribute
%A better performance for 2-layer DGLSTM-CRF compared to one layer only implies the effectiveness of the grandchild dependencies (e.g., ``(\textit{prep, pobj})''). 

\begin{table}
	\centering
	\resizebox{\linewidth}{!}{
		\begin{tabular}{lccc}
			\toprule
			\textbf{Model}& \textbf{Prec.} & \textbf{Rec.} & \textbf{F}$_\mathbf{1}$ \\
			\midrule
			
			%			\multicolumn{3}{l}{\textbf{Pretrained Word Embedding}} &\\[1mm]
			\citet{chiu2016named}  &86.04 & 86.53 & 86.28\\
			\citet{li2017leveraging}   & 88.00 & 86.50 & 87.21 \\
			\citet{ghaddar2018robust} &- &- & 87.95 \\
			\citet{strubell2017fast} &-& -&   86.84 \\
			\cdashlinelr{1-4}
			BiLSTM-CRF ($L=0$) &82.03 & 80.78&81.40  \\
			BiLSTM-CRF ($L=1$) & 87.21 & 86.93 & 87.07 \\
			BiLSTM-CRF ($L=2$)  & 87.89 & 87.68 & 87.78 \\
			BiLSTM-CRF ($L=3$)  &  87.81& 87.50 & 87.65 \\
			%			DGGCN-CRF ($J=1$)  & 87.56 & 87.82 & 87.69 \\
			BiLSTM-GCN-CRF  &88.30 & 88.06 & 88.18 \\
			\cdashlinelr{1-4}
			DGLSTM-CRF ($L=0$) &85.31  & 82.19 & 84.09 \\
			DGLSTM-CRF ($L=1$) & \textbf{88.78} & 87.29 & 88.03 \\
			DGLSTM-CRF ($L=2$)& 88.53 & \textbf{88.50} & {\bf 88.52} \\
			DGLSTM-CRF ($L=3$)& 87.59 & 88.93 & 88.25 \\[1mm]
			
			\multicolumn{3}{l}{\textbf{Contextualized Word Representation}} &\\
			\citet{akbik2018coling} (Flair) & - & - & 89.30 \\
			\cdashlinelr{1-4}
			BiLSTM-CRF ($L=0$) + {\footnotesize ELMo} & 85.44& 84.41 & 84.92 \\
			BiLSTM-CRF ($L=1$) + {\footnotesize ELMo} & 89.14 & 88.59 & 88.87 \\
			BiLSTM-CRF  ($L=2$) + {\footnotesize ELMo}  & 88.25 & 89.71 & 88.98 \\
			BiLSTM-CRF ($L=3$) + {\footnotesize ELMo} & 88.03 & 89.04 & 88.53 \\
			
			BiLSTM-GCN-CRF +  {\footnotesize ELMo}   & 89.40 & 89.71 & 89.55 \\
			\cdashlinelr{1-4}
			DGLSTM-CRF ($L=0$) + {\footnotesize ELMo} &86.87  &85.12  &85.99 \\
			DGLSTM-CRF ($L=1$) + {\footnotesize ELMo} & 89.40 & 89.96 & 89.68 \\
			DGLSTM-CRF ($L=2$) + {\footnotesize ELMo} & \textbf{89.59} & \textbf{90.17} & {\bf 89.88} \\
			DGLSTM-CRF ($L=3$) + {\footnotesize ELMo} & 89.43 & 90.15 & 89.79 \\
			%			DGGCN-CRF (2-layer) + ELMo  & 88.80 & 89.74 & 89.27 \\
			\bottomrule
		\end{tabular}
	}
%	\vspace*{-3mm}
	\caption{Performance comparison on the OntoNotes 5.0 English dataset.}
%	\vspace*{-4mm}
	\label{tab:mainresults}
\end{table}

%To further investigate if the effect of dependency would be diminished by the contextualized word representations, we compare the performance of all models with ELMo~\cite{peters2018deep} representations (Table \ref{tab:mainresults} bottom). 
We further compare the performance of all models with ELMo~\cite{peters2018deep} representations to investigate whether the effect of dependency would be diminished by the contextualized word representations. 
With $L$ = $0$, the ELMo representations largely improve the performance of BiLSTM-CRF compared to the BiLSTM-CRF model with word embeddings only but is still 1 point below our DGLSTM-CRF model. 
The 2-layer DGLSTM-CRF model outperforms the best BilSTM-CRF baseline with 0.9 points in F$_1$ ($p<0.001$).
%The improvements apply to both precision and recall. 
%We attribute the precision gain to the long-distance dependency edges and the recall gain to the dependency label information. 
Empirically, we found that among the entities that are correctly predicted by DGLSTM-CRF but wrongly predicted by BiLSTM-CRF, 47\% of them are with length more than 2. 
Our finding shows the 2-layer DGLSTM-CRF model is able to accurately recognize long entities, which can lead to a higher precision.  
%These entities have either a length more than 4 or a dependency parent far from them. 
In addition, 51.9\% of the entities that are correctly retrieved by DGLSTM-CRF have the dependency relations ``\textit{pobj}'', ``\textit{nn}'' and ``\textit{nsubj}'', which have strong correlations with certain named entity types (Figure \ref{fig:depstat}).  
Such a result demonstrates the effectiveness of dependency relations in improving the recall of NER.
%The improvements on recall also demonstrate the effectiveness of dependency relations in retrieving more entities. 

\begin{table}
	\centering
	\resizebox{\linewidth}{!}{
		\begin{tabular}{lccc}
			\toprule
			\textbf{Model}& \textbf{Prec.} & \textbf{Rec.} & \textbf{F}$_\mathbf{1}$ \\
			\midrule
			\citet{pradhan2013towards}  &78.20 & 66.45 & 71.85\\ 
%			\citet{nosirova2019effective}$\dagger$ & -& -& 72.40\\
			Lattice LSTM (\textcolor{darkblue}{Z\&Y, 2018}) & 76.34& 77.01& 76.67 \\
			\cdashlinelr{1-4}
			BiLSTM-CRF ($L=0$) & 76.67& 67.79& 71.95  \\
			BiLSTM-CRF ($L=1$)  & \textbf{78.45} & 74.59 & 76.47 \\
			BiLSTM-CRF ($L=2$)  & 77.94 & 75.33 & 76.61 \\
			BiLSTM-CRF ($L=3$)  & 76.17 & 75.23 & 75.70 \\
			
			BiLSTM-GCN-CRF  & 76.35 & 75.89 & 76.12 \\
			\cdashlinelr{1-4}
				DGLSTM-CRF ($L=0$) & 76.91&70.65& 73.65 \\
			DGLSTM-CRF ($L=1$)   & 77.79 & 75.29 & 76.52 \\
			DGLSTM-CRF ($L=2$)   & 77.40 & \textbf{77.41} & \textbf{77.40} \\
			DGLSTM-CRF ($L=3$)&77.01  &74.90  &75.94 \\[1mm]
%			BiLS ($L=1$) & 77.59 & 73.67 & 75.58 \\
	\multicolumn{3}{l}{\textbf{Contextualized Word Representation}} &\\
			BiLSTM-CRF ($L=0$) + {\footnotesize ELMo} & 75.20 & 73.39 & 74.28 \\
			BiLSTM-CRF ($L=1$) + {\footnotesize ELMo}  & \textbf{79.20} & 79.21 & 79.20 \\
			BiLSTM-CRF($L=2$) + {\footnotesize ELMo}  & 78.49 & 79.44 & 78.96 \\
			BiLSTM-CRF ($L=3$) + {\footnotesize ELMo} &78.54  &79.76  &79.14 \\
					
			BiLSTM-GCN-CRF + {\footnotesize ELMo}   & 78.71 & 79.29 & 79.00 \\
			\cdashlinelr{1-4}
			DGLSTM-CRF ($L=0$) + {\footnotesize ELMo} & 76.27 & 74.61 &75.43 \\
			DGLSTM-CRF ($L=1$) + {\footnotesize ELMo}   &78.91&80.22& 79.56\\
			DGLSTM-CRF ($L=2$) + {\footnotesize ELMo}   & 78.86 & \textbf{81.00} & {\bf 79.92} \\
			DGLSTM-CRF ($L=3$) + {\footnotesize ELMo} & 79.30 &79.86  &79.58  \\
			\bottomrule
		\end{tabular}
	}
%	\vspace*{-2.5mm}
	\caption{Performance comparison on the OntoNotes 5.0 Chinese Dataset.}
%	\vspace*{-4mm}
	\label{tab:mainresults_chinese}
\end{table}

\paragraph{OntoNotes Chinese}
Table \ref{tab:mainresults_chinese} shows the performance comparison on the Chinese datasets. 
\begin{table*}
	\centering
%	\resizebox{0.8}{!}{
			\scalebox{0.8}{
		\begin{tabular}{lcccccc}
			\toprule
			\multirow{2}{*}{\textbf{Model}}&  \multicolumn{3}{c}{\textbf{Catalan}}& \multicolumn{3}{c}{\textbf{Spanish}}\\[-1mm]
			\cmidrule(lr){2-4} \cmidrule(lr){5-7} 
			& \textbf{Prec.} & \textbf{Rec.} & \textbf{F}$_\mathbf{1}$& \textbf{Prec.} & \textbf{Rec.} & \textbf{F}$_\mathbf{1}$ \\
			\midrule
			%			\multicolumn{9}{l}{\textbf{Pretrained Word Embedding}} &\\[1mm]
			BiLSTM-CRF ($L=0$) & 65.91 & 49.90 & 56.80 & 65.97&52.63&58.55  \\
			BiLSTM-CRF ($L=1$) & 76.83 & 63.47 & 69.51& 78.33 & 69.89 & 73.87 \\
			BiLSTM-CRF ($L=2$)  & 73.79 & 67.63 & 70.58 & 77.73 & 70.91 & 74.16 \\
			BiLSTM-CRF ($L=3$) & 74.75 & 67.35 & 70.86 & 76.41& 72.95 & 74.64 \\
			%			DGGCN-CRF ($L=1$) &83.38 & 84.63 & 84.00& 79.55 & 72.71 & 75.95 & 	84.82 & 78.53 & 81.55 \\
			BiLSTM-GCN-CRF    & 81.25 & 75.22 & 78.12&84.10 & 79.88 & 81.93\\
			\cdashlinelr{1-7}
			DGLSTM-CRF ($L=0$) & 73.42 & 	61.79 & 67.10& 74.90& 61.21 & 67.38\\ 
			DGLSTM-CRF ($L=1$)   & 81.87 & 79.28 & 80.55 & 83.21 & 81.19 & 82.19 \\
			DGLSTM-CRF ($L=2$)   & \textbf{83.35} & \textbf{80.00} & \textbf{81.64} & 84.05 & 82.90 &83.47\\
			DGLSTM-CRF ($L=3$) & 81.87 & 80.21 & 81.03 & \textbf{84.12} & \textbf{83.45} & \textbf{83.78}\\ 
			\multicolumn{6}{l}{\textbf{Contextualized Word Representation}} &\\[1mm]
			BiLSTM-CRF ($L=0$)  + {\footnotesize ELMo} & 67.53 & 64.47 & 65.96& 73.16 & 69.01 & 71.03\\
			BiLSTM-CRF ($L=1$) + {\footnotesize ELMo}   &77.85 & 76.22 & 77.03& 81.72 & 79.09 & 80.38 \\
			BiLSTM-CRF($L=2$) + {\footnotesize ELMo}  & 78.61 & 78.32 & 78.46 & 80.89 & 80.30 & 80.59\\
			BiLSTM-CRF($L=3$) + {\footnotesize ELMo}  & 79.11 & 77.32 & 78.21 & 80.48 & 79.45 & 79.96\\
			BiLSTM-GCN-CRF  + {\footnotesize ELMo}    &83.68 & 83.16 & 83.42 & 85.31 & 85.19 & 85.25\\
			%			DGGCN-CRF ($L=2$) + {\footnotesize ELMo}   & 83.92 & 84.86 & 84.38 & 83.61 & 83.44 & 83.52 &85.15 & 85.46 & 85.30\\
			\cdashlinelr{1-7}
			DGLSTM-CRF ($L=0$) + {\footnotesize ELMo}   & 70.87 & 65.81 & 68.25 &  75.96& 72.52 & 74.20\\
			DGLSTM-CRF ($L=1$) + {\footnotesize ELMo}    & 82.29 & 82.37 & 82.33 & 84.05 & 84.77 & 84.41\\
			DGLSTM-CRF ($L=2$) + {\footnotesize ELMo}    & \textbf{84.71} & 83.75 & \textbf{84.22}& \textbf{87.79} & \textbf{87.33} & \textbf{87.56} \\
			DGLSTM-CRF ($L=3$)  + {\footnotesize ELMo}& 84.50 & \textbf{83.92} & \textbf{84.21} & 86.74 & 86.57 & 86.66 \\
			\bottomrule
		\end{tabular}
	}
%	\vspace*{-2.5mm}
	\caption{Results on the SemEval-2010 Task 1 datasets.}
%	\vspace*{-3mm}
	\label{tab:semevalresult}
\end{table*}
We compare our models against the state-of-the-art NER model on this dataset, Lattice LSTM~\cite{zhang2018chinese}\footnote{We run their code on the OntoNotes 5.0 Chinese dataset.}. 
Our implementation of the strong BiLSTM-CRF baseline achieves comparable performance against the Lattice LSTM. 
Similar to the English dataset, our model with $L$ = $0$ significantly improves the performance compared to the BiLSTM-CRF ($L$ = $0$) model. 
Our DGLSTM-CRF model achieves the best performance with $L$ = $2$ and is consistently better ($p<0.02$) than the strong BiLSTM-CRF baselines. 
%Although the 2-layer DGLSTM-CRF model consistently achieves the best results in F$_1$ ($p<0.05$) compared to our strong BiLSTM-CRF baselines, the improvements are not as significant as in the English dataset. 
%The DGGCN-CRF obtains only comparable performance against the baselines. 
As we can see from the table, the improvements of the DGLSTM-CRF model mainly come from recall ($p<0.001$) compared to the BiLSTM model, especially in the scenario with word embeddings only. 
%Such improvements leverage the ability of dependency relations in helping to predict more entities. 
Empirically, we also found that those correctly retrieved entities of the DGLSTM-CRF (compared against the baseline) mostly correlate with the following dependency relations: ``\textit{nn}'', ``\textit{nsubj}'', ``\textit{nummod}''.
However, DGLSTM-CRF achieves lower precisions against BiLSTM-CRF, which indicates that the DGLSTM-CRF model makes more false-positive predictions. 
%The reason could be that there are some dependency relations that strongly correlates entity words also correlate to non-entity words. 
%As such, the model would tend to predict the words  with those dependency relations  as entities while these words are actually non-entity. 
The reason could be the relatively lower ratio of ST(\%)\footnote{Percentage of entities that can form a subtree.} as shown in Table \ref{tab:datastat}, which means some of the entities do not form subtrees under the complete dependency trees. 
In such a scenario, the model may not correctly identify the boundary of the entities, which results in lower precision. 
%We found that the number of correctly predicted entities by DGLSTM-CRF is actually larger than the one by BiLSTM-CRF. 
%However, the DGLSTM-CRF predicted more entities and achieved lower precision, which might be caused by some dependency labels that intend to predict more entities. 

%The contextualized word representation is able to significantly improve the recall 
%Only recall improved. Chinese low agreement rate 
%low parsing performance. lower than English

\paragraph{SemEval-2010} 

%Table \ref{tab:semevalresult} shows the results of our models on the SemEval-2010 Task 1 datasets. 
%%The results marked in bold are significantly better than the best performing BiLSTM-CRF baseline. 
%As the English dataset is a subset of the OntoNotes English corpus, we obtain similar performance as in the OntoNotes 5.0 English dataset where the DGLSTM-CRF achieves the best performance ($p<0.05$) in precision, recall and F$_1$. 
%The improvement are not as significant as in the OntoNotes English dataset. 
%The reason is similar to the Chinese dataset where we have a relatively low ST(\%) in this dataset. 

Table \ref{tab:semevalresult} shows the results of our models on the SemEval-2010 Task 1 datasets.
Overall, we observe substantial improvements of the DGLSTM-CRF on the Catalan and Spanish datasets (with $p<0.001$ marked in bold against the best performing BiLSTM-CRF baseline), especially for DGLSTM-CRF with ELMo and $L$ larger than 1.  
%Such improvements demonstrate the effectiveness of the DGLSTM-CRF in encoding the dependency trees for NER. 
With word embeddings, the best DGLSTM-CRF model outperforms the best performing BiLSTM-CRF baseline with more than 10 and 9 points in  F$_1$  on the Catalan and Spanish datasets, respectively. 
The BiLSTM-GCN-CRF model also performs much better than the BiLSTM-CRF baselines but is worse than the DGLSTM-CRF model with $L\geq 2$. 
%The improvements over the BiLSTM-GCN-CRF further shows the importance of 
%the DGLSTM-CRF model has more than 10\% F$_1$ improvement over the BiLSTM-CRF model on the Catalan dataset and about 9\% improvement on the Spanish dataset. 
Both precision and recall significantly improve with a large margin compared to the best performing BiLSTM-CRF, especially for the recall (with more than 10 points improvement) on these two datasets. 
With ELMo, the best performing DGLSTM-CRF model outperforms the BiLSTM-CRF baseline with about 6 and 7 points in F$_1$ on these two datasets, respectively. 
The substantial improvements show that the structural dependency information is extremely helpful for these two datasets. 
%the performance of BiLSTM-CRF improves for more than 6 points in  F$_1$. 
%On the other hand, the F$_1$ of the DGLSTM-CRF with 1 layer also increases for these two languages and still significantly better than the baselines.

%Such improvements not only apply to recall but precision as well. 
%The absolute improvements is even larger than the one we observe in the OntoNotes English datasets. 
 
%The improvement of DGGCN-CRF is not as significant as DGLSTM-CRF, showing that the DGLSTM-CRF is more robust across different languages and the ability of LSTM that captures the contextualized information is important for the NER task. 

With ELMo representations, we observe about 2 and 3 points improvements in F$_1$  compared with the 1-layer DGLSTM-CRF model on these two datasets, respectively. 
%We further compare the results between the 1-layer DGLSTM-CRF and 2-layer DGLSTM-CRF (with ELMo) models to investigate their differences in the Catalan and Spanish datasets. 
%Empirically, the 2-layer DGLSTM-CRF model improves the the performance of all entity types (e.g., \textsc{person}) compared to the 1-layer DGLSTM-CRF model. 
Empirically, more than 50\% of the entities that are correctly predicted by the 2-layer model but not the 1-layer model are with length larger than 2.
%Empirically, the performance of all entity types (e.g., \textsc{person}) is improved by 
%has significant improvement with the 2-layer model. 
%Among the entities that are correctly recognized by 2-layer model but not 1-layer model, more than 50\% of them are entities with length more than 2. 
%We found that, among the entities correctly predicted  by the 2-layer model but not the 1-layer , more than 50\% of them with length larger than 2. 
Also, most of these entities contain the grandchild dependencies ``(\textit{sn, sn})'' and ``(\textit{spec, sn})'' where \textit{sn} represents noun phrase and \textit{spec} represents specifier (e.g., determiner, quantifier) in both datasets. 
Such a finding shows that the 2-layer model is able to capture the interactions given by the grandchild dependencies. 

\subsection{Additional Experiments}
%We conduct additional experiments on the CoNLL-2003 English dataset and ablation study on the OntoNotes English dataset. 
%
\paragraph{CoNLL-2003 English}
Table \ref{tab:resconll} shows the performance on the CoNLL-2003 English dataset. 
The dependencies are predicted from Spacy~\cite{spacy2}. 
With the contextualized word representations, DGLSTM-CRF outperforms BiLSTM-CRF with 0.2 points in F$_1$ ($p<0.09$). 
%The improvement is not significant due to the relatively low ST ratio as shown Table \ref{tab:datastat}. 
The improvement is not significant due to the relatively lower equality of the dependency trees. 
To further study the effect of the dependencies, we modified the predicted dependencies to ensure each entity form a subtree in the complete dataset. 
Such modification improves the F$_1$ to 92.7, which is significantly better ($p<0.05$) than the BiLSTM-CRF.

%Similar to the SemEval English and OntoNotes Chinese datasets, the moderate  improvements are caused by the relatively low ST ratio as shown in Table \ref{tab:datastat}. 
\begin{table}[t!]
	\centering
	\resizebox{1.0\linewidth}{!}{
% 	\scalebox{0.6}{
		\begin{tabular}{lccc}
			\toprule
			%			\multirow{2}{*}{\bf Model}& \multicolumn{3}{c}{\textbf{Test}} \\
			\textbf{Model} & \textbf{Prec.} & \textbf{Rec.} &   \textbf{F}$_\mathbf{1}$\\
			\midrule
			\citet{peters2018deep}~ ELMo  &  -& - &92.2 \\
			\cdashlinelr{1-4}
			BiLSTM-CRF  + {\footnotesize ELMo ($L=2$)}  &  92.1 & 92.3 &    92.2 \\
			DGLSTM-CRF + {\footnotesize ELMo ($L=2$)} & 92.2&92.5 &92.4\\
			\bottomrule
		\end{tabular}
	}
	\vspace{-2.5mm}
	\caption{Performance on the CoNLL-2003 English dataset.}
	\vspace{-1.5mm}
	\label{tab:resconll}
\end{table}

\begin{table}
	\centering
	\resizebox{1.0\linewidth}{!}{
		%	\scalebox{0.8}{
		\begin{tabular}{lcccccc}
			\toprule
			\multirow{2}{*}{\textbf{Model}}&  \multicolumn{3}{c}{\textbf{Catalan}}& \multicolumn{3}{c}{\textbf{Spanish}}\\[-1mm]
			\cmidrule(lr){2-4} \cmidrule(lr){5-7} 
			& \textbf{Prec.} & \textbf{Rec.} & \textbf{F}$_\mathbf{1}$& \textbf{Prec.} & \textbf{Rec.} & \textbf{F}$_\mathbf{1}$ \\
			\midrule
			%			\multicolumn{9}{l}{\textbf{Pretrained Word Embedding}} &\\[1mm]
			BiLSTM-CRF ($L=1$) & 47.88 & 18.59 & 26.78 & 40.77 & 19.01 & 25.93 \\
			DGLSTM-CRF ($L=1$) & 47.71 & 31.55 & 37.98&  49.39 & 31.91 & 38.77\\
			 ~~-- with gold dependency & \textbf{52.13} & \textbf{33.26} & \textbf{40.61} & \textbf{52.14} & \textbf{35.59} & \textbf{42.30} \\
			\bottomrule
		\end{tabular}
	}
%	\vspace*{-2.5mm}
	\caption{Low-resource NER performance on the SemEval-2010 Task 1 datasets.}
%	\vspace*{-4mm}
	\label{tab:lowresources}
\end{table}

\paragraph{Low-Resource NER}
Following \citet{cotterell2017low}, we emulate truly low-resource condition with 100 sentences for training.
We assume that the contextualized word representations are not available and dependencies are predicted. 
Table \ref{tab:lowresources} shows the NER performance on the SemEval-2010 Task 1 datasets under the low-resource setting. 
With limited amount of training data, BiLSTM-CRF suffers from low recall and the DGLSTM-CRF largely improves it on these two datasets. 
Using gold dependencies further significantly improves the precision and recall. 
%Further gold dependency annotations can 
%We can see that the DGLSTM-CRF largely improves the recall 

% \begin{table}[t!]
%     \centering
%     \resizebox{1.0\linewidth}{!}{
%     \begin{tabular}{lcccc}
%         \toprule
%          & English & Chinese & Catalan & Spanish \\
%          \midrule
%         UAS &96.0 & 91.4& 95.2& 95.3\\
%         LAS & 94.9&89.3 &93.3 & 93.4\\
%         \bottomrule
%     \end{tabular}
%     }
%     \caption{The performance of dependency parser on all languages. UAS and LAS represent the unlabeled and labeled attachment scores, respectively.}
%     \label{tab:dep_result}
% \end{table}

% \begin{table}[t!]
% 	\centering
% 	\resizebox{1\linewidth}{!}{
% 		\begin{tabular}{lccccccc}
% 			\toprule
% 			\multirow{2}{*}{\textbf{Dataset}}&  \multicolumn{3}{c}{\textbf{Gold Dependency}}& \multicolumn{4}{c}{\textbf{Predicted Dependency}}\\[-1mm]
% 			\cmidrule(lr){2-4} \cmidrule(lr){5-8} 
% 			& \textbf{Prec.} & \textbf{Rec.} & \textbf{F}$_\mathbf{1}$ & \textbf{LAS}& \textbf{Prec.} & \textbf{Rec.} & \textbf{F}$_\mathbf{1}$ \\
% 			\midrule
		
% 			English & 89.59 & 90.17 & 89.88 & 94.9 & 89.62 & 89.67 & 89.64\\
% 			Chinese &  78.86 & 81.00 &79.92 & 89.3 & 78.51 & 80.70 & 79.59\\
% 			Catalan & 84.71 & 83.75 & 84.22 & 93.3 & 83.28 & 81.48 & 82.37\\
% 			Spanish &  87.79& 87.33 & 87.56 & 93.4 &  83.28 & 84.57 & 83.92\\
% 			\bottomrule
% 		\end{tabular}
% 	}
% %	\vspace*{-3mm}
% 	\caption{Performance comparision.}
% %	\vspace*{-4mm}
% 	\label{tab:dep_result}
% \end{table}

\begin{table}[t!]
	\centering
	\resizebox{1\linewidth}{!}{
		\begin{tabular}{lcccc}
			\toprule
			& \textbf{English} & \textbf{Chinese} & \textbf{Catalan} & \textbf{Spanish} \\
			\midrule
		    BiLSTM-CRF & 88.98 & 79.20 & 78.46 & 80.59 \\
		    \midrule
		    (Dependency LAS)$\dagger$ & (94.89) & (89.28) & (93.25) & (93.35)\\
		    DGLSTM-CRF (Predicted) & \textbf{89.64} & \textbf{79.59} & \textbf{82.37} & \textbf{83.92} \\
		    Improvement $\Delta$ & +0.66 & +0.39 & +3.91 & +3.33\\
		    \midrule \midrule
		    DGLSTM-CRF (Gold) & 89.88 & 79.92 & 84.22 & 87.56 \\ 
			\bottomrule
		\end{tabular}
	}
%	\vspace*{-3mm}
	\caption{F$_1$ performance of DGLSTM-CRF with predicted dependencies against the best performing BiLSTM-CRF. $\dagger$: LAS is label attachment score which is the metric for dependency evaluation.}
%	\vspace*{-4mm}
	\label{tab:dep_result}
\end{table}

\begin{table}[t!]
	\centering
	\resizebox{1.0\linewidth}{!}{
		\begin{tabular}{lccc}
			\toprule
%			\multirow{2}{*}{\bf Model}& \multicolumn{3}{c}{\textbf{Test}} \\
			\textbf{Model}&  \textbf{Prec.} & \textbf{Rec.} & \textbf{F}$_\mathbf{1}$\\
			\midrule
			BiLSTM-CRF + {\footnotesize ELMo ($L=2$)}  & 89.14& 88.59 & 88.87 \\
			\cdashlinelr{1-4}
			DGLSTM-CRF  + {\footnotesize ELMo ($L=2$)}  & \textbf{89.59}& \textbf{90.17} & \textbf{89.88} \\
			~~~--$g(\cdot) = $ self connection & 89.17 &90.08  &89.62   \\
			~~~--$g(\cdot) = $ Concatenation & 89.43 & 90.09 & 89.76  \\
			~~~--$g(\cdot) = $ Addition & 89.24&89.78 & 89.50\\
			~~~--w/o dependency relation & 88.92 & 89.99 & 89.46 \\
%			~~--with Universal dependency & 87.33 &89.33 & 88.32 \\
% 			~~~--with predicted dependency &89.62 & 89.67 & 89.64 \\
			\bottomrule
		\end{tabular}
	}
%	\vspace{-2.5mm}
	\caption{Ablation study of the DGLSTM-CRF model on the OntoNotes English dataset.}
%	\vspace{-3mm}
	\label{tab:ablation}
\end{table}

\paragraph{Effect of Dependency Quality}
% In practice, it is laborious to obtain parse tree annotations for a sentence. 
To evaluate how the quality of dependency trees affect the performance, we train a state-of-the-art dependency parser~\cite{dozat2017deep} using our training set and make prediction on the development/test set. 
We implemented the dependency parser using the AllenNLP package~\cite{Gardner2017AllenNLP}.
Table \ref{tab:dep_result} shows the performance (LAS) of the dependency parser on four languages (i.e., OntoNotes English, OntoNotes Chinese, Catalan and Spanish) and the performance of DGLSTM-CRF against the best performing BiLSTM-CRF with ELMo.
% As we can see, the dependency performance on Chinese is relatively worse than the performance on other languages. 
% We attribute this performance to 
DGLSTM-CRF even with predicted dependencies is able to consistently outperform the BiLSTM-CRF on four languages. 
However, the performance is still worse than the DGLSTM-CRF with gold dependencies, especially on the Catalan and Spanish.
Such results suggest that it is essential to have high-quality dependency annotations available for the proposed model.
% However, the dependency parsers are not perfect enough to obtain completely accurate parse trees.
% Such predicted dependencies make the performance of DGLSTM-CRF worse than the DGLSTM-CRF with gold dependencies, especially on the Catalan and Spanish languages where the dependencies are more effective as shown in Table \ref{tab:semevalresult}.
% We can see the performance on Chinese is not good.

\paragraph{Ablation Study}
%We conduct ablation study in Table \ref{tab:ablation} on the best performing 2-layer DGLSTM-CRF model. 
Table \ref{tab:ablation} shows the ablation study of the 2-layer DGLSTM-CRF model on the OntoNotes English dataset.
With self connection as interaction function, the F$_1$ drops 0.3 points.
The model achieves comparable performance with concatenation as interaction function but  F$_1$ drops about 0.4 points with the addition interaction function.  
We believe that the addition potentially leads to certain information loss.
Without the dependency relation embedding $\vec{v}_r$ in the input representation, the F$_1$ drops about 0.4 points.
%We also tried with the Universal dependency representation\footnote{https://universaldependencies.org/} converted from the Stanford CoreNLP Toolkit. 
%The performance on development set increase about 0.2\% in F$_1$ but it does not generalize to the test set. 
%The performance on development set increase about 0.2\% in F$_1$. 
% Using predicted dependency from Spacy~\cite{spacy2} degrade the performance for about 0.6 points, which shows that the quality of dependency is important for achieving high performance. 
% Overall, the performance with predicted dependencies is still better than the best performing BiLSTM-CRF. 

\begin{table}[t!]
	\centering
	\resizebox{1.0\linewidth}{!}{
	\begin{tabular}{clcccccc}
		\toprule
		\multirow{2}{*}{\textbf{Dataset}}& \multirow{2}{*}{\textbf{Model}} & \multicolumn{6}{c}{\textbf{Entity Length}} \\
		 & & \textbf{1} & \textbf{2} & \textbf{3} & \textbf{4} & \textbf{5}  & $\mathbf{\geq}$\textbf{6}  \\
		 \midrule
		 \multirow{2}{*}{\textbf{English}} & BiLSTM-CRF & \textbf{91.8} & 88.5 & 83.4 & 84.0 & 75.4 & 76.0 \\
		 & DGLSTM-CRF & \textbf{91.8} &  \textbf{90.1} &  \textbf{85.4}&  \textbf{87.0} &  \textbf{80.8} &  \textbf{78.7} \\ 
		 \midrule
		 \multirow{2}{*}{\textbf{Chinese}} & BiLSTM-CRF &81.2 & 74.3 & \textbf{73.1} & 62.8 & \textbf{70.3} & \textbf{57.5} \\
		 & DGLSTM-CRF & \textbf{82.2}& \textbf{75.5} & 71.8 & \textbf{64.1} & 58.5 & 41.1\\
		 \midrule
		 \multirow{2}{*}{\textbf{Catalan}} & BiLSTM-CRF &80.5 & 81.0 & 75.8 & 56.1 & 45.0 & 38.4\\
		 & DGLSTM-CRF & \textbf{85.4} & \textbf{85.1} & \textbf{84.1} & \textbf{78.9} & \textbf{60.9} & \textbf{59.3}\\
		 \midrule
		 \multirow{2}{*}{\textbf{Spanish}} & BiLSTM-CRF & 84.2 & 81.1 & 81.0 & 53.3 & 53.3 & 37.1\\
		 & DGLSTM-CRF & \textbf{89.3} & \textbf{87.4} & \textbf{90.8} & \textbf{74.1} & \textbf{67.7} & \textbf{64.4}\\
		 
		 \bottomrule
	\end{tabular}
	}
%	\vspace{-2.5mm}
	\caption{Performance of entities with different lengths on the four datasets: OntoNotes (English), OntoNotes Chinese, Catalan and Spanish.}
%	\vspace{-3mm}
	\label{tab:reslength}
\end{table}

\section{Analysis}
\label{sec:analysis}
% We conduct analysis with the 2-layer DGLSTM-CRF on the OntoNotes English dataset.
\subsection{Effectiveness of Dependency Relations}
\begin{figure}[t!]
	\centering
	\scalebox{1.0}{
	\includegraphics[width=3.2in]{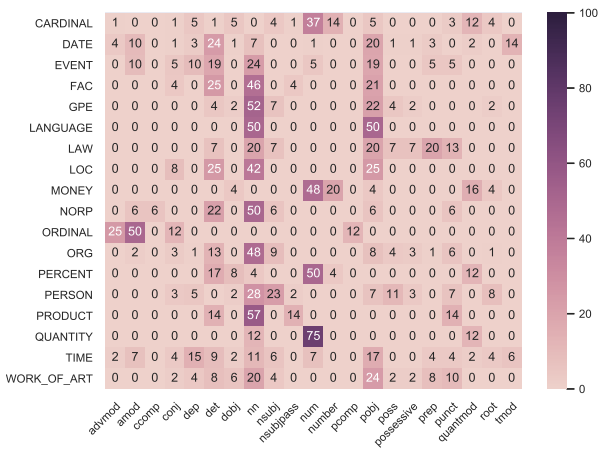}
}
%	\vspace{-8mm}
	\caption{Correlations between the correctly predicted entities and the dependency relations.}
	\label{fig:deprelana}
%	\vspace{-2mm}
\end{figure}

To demonstrate whether the model benefits from the dependency relations, we first select the entities that are correctly predicted by the 2-layer DGLSTM-CRF model but not by the best performing baseline 2-layer BiLSTM-CRF on the OntoNotes English dataset. 
We draw the heatmap in Figure \ref{fig:deprelana} based on these entities.  
Comparing Figure \ref{fig:depstat} and \ref{fig:deprelana}, we can see that they are similar in terms of the density. 
Both of them show consistent relationships between the entity types and the dependency relations. 
The comparison shows that the improvements partially result from the effect of dependency relations. 
We also found from our model's predictions that some entity types have strong correlations with the relation pairs on grandchild dependencies\footnote{The corresponding heatmap visualization is provided in supplementary material.}. 

\subsection{Entity with Different Lengths}
Table \ref{tab:reslength} shows the performance comparison with different entity lengths on all datasets. 
As mentioned earlier, the dependencies as well as the grandchild relations allow our models to capture the long-distance interactions between the words. 
% The performance of NER can benefit from such long-distance interactions, especially for long entities. 
As shown in the table, the performance of entities with lengths more than 1 consistently improves with DGLSTM-CRF for all languages except Chinese.
As we pointed out in the dataset statistics (Table \ref{tab:datastat}),  the number of entities that form subtrees in OntoNotes Chinese is relatively smaller compared to other datasets.
The performance gain is more significant for entities with longer length on the other three languages. 
We found that, among the improvements of entities with length larger than 2 in English, 85\% of them have long-distance dependencies and 30\% of them have grandchild dependencies within the entity boundary. 
The analysis shows that our model that exploits the dependency tree structures is helpful for recognizing long entities.
%Such result statistics show the DGLSTM-CRF is able to capture the long-distance interactions. 
%

\section{Conclusions and Future Work}
Motivated by the relationships between the dependency trees and named entities, we propose a dependency-guided LSTM-CRF model to encode the complete dependency tree and capture such relationships for the NER task. 
Through extensive experiments on several datasets, we demonstrate the effectiveness of the proposed model in improving the NER performance. 
%Such improvements also generalize to a low-resource setting. 
Our analysis shows that NER benefits from the dependency relations and long-distance dependencies, which are able to capture the non-local interactions between the words. 
%Our quantified analysis show that NER benefits from the dependency relations and the long-distance interactions between the words. 

As statistics shows that most of the entities form subtrees under the dependency trees, future work includes building a model for joint NER and dependency parsing which regards each entity as a single unit in a dependency tree. 

\section*{Ackowledgements}
We would like to thank the anonymous reviewers for their constructive comments on this work. 
We would also like to thank Zhijiang Guo and Yan Zhang for the fruitul discussion. 
This work is supported by Singapore Ministry of Education Academic Research Fund (AcRF) Tier 2 Project MOE2017-T2-1-156.

\bibliography{emnlp_depner}
\bibliographystyle{acl_natbib}

\newpage

\appendix

\section{Baseline Systems}
We implemented the BiLSTM-CRF~\cite{lample2016neural} and BiLSTM-GCN-CRF models based on the contextualized GCN implementation by \citet{zhang2018graph}. 
The implementation of BiLSTM-CRF is exactly same as \citet{lample2016neural}. 
We presents the neural architecture for the BiLSTM-GCN-CRF model. 
\subsection{BiLSTM-GCN-CRF}
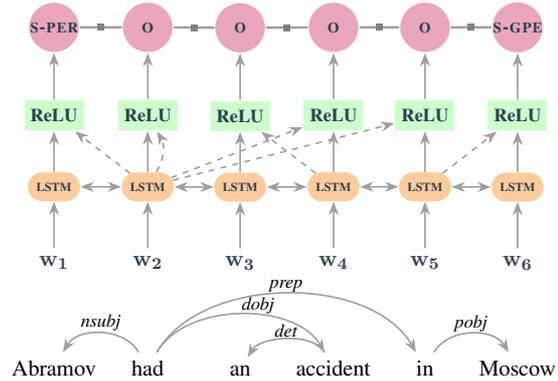
\begin{figure}[t!]
	\centering
	\adjustbox{max width=1\linewidth}{
		\begin{tikzpicture}[node distance=8.0mm and 10mm, >=Stealth, 
		sentity/.style={draw=none, circle, minimum height=8.5mm, minimum width=8.5mm,line width=1pt, inner sep=0pt, fill=purple!35},
		lstm/.style={draw=none, minimum height=5mm, rounded rectangle, fill=puffin!50, minimum width=11.5mm, label={center:\tiny \textcolor{fontgray}{\bf LSTM}}},
		%		, blur shadow={shadow blur steps=5}
		plus/.style={draw=none, minimum height=3mm, circle, fill={rgb:red,0;green,210;blue,211}, label={center:\footnotesize \textcolor{white}{+}}},
		gfunc/.style={draw=none, minimum height=5mm, rectangle, fill=green!20, label={center:\footnotesize \textcolor{fontgray}{\bf ReLU}}, minimum width=10mm, line width = 1.5pt},
		emb/.style={draw=none, minimum height=3mm, rounded rectangle, fill=none, minimum width=12mm, text=fontgray},
		%		cyan!40
		chainLine/.style={line width=0.8pt,->, color=mygray},
		%	background rectangle/.style={fill=olive!45}, show background rectangle, 
		%		line width=1.3pt
		]

		\matrix (s) [matrix of nodes, nodes in empty cells, execute at empty cell=\node{\strut};]
		{
			Abramov & [2ex]had & [5ex]an & [3ex]accident   &   [3ex] in  & [3ex] Moscow\\
			%		 \textbf{\textsc{date}}   &   &   & \textsc{o}     & \textsc{o}   & \textsc{o}  & \textsc{o}& \textsc{event} &  \\
		};
		
		\draw [chainLine, ->] (s-1-2) to [out=120,in=60, looseness=1] node[above, yshift=-1mm, color=black]{\footnotesize\it nsubj} (s-1-1);
		\draw [chainLine, ->] (s-1-4) to [out=120,in=60, looseness=0.8] node[above, yshift=-1mm, color=black]{\footnotesize\it det} (s-1-3);
		\draw [chainLine, ->] (s-1-2) to [out=60,in=120, looseness=0.9] node[above, yshift=-1.2mm, xshift=3mm, color=black]{\footnotesize\it dobj} (s-1-4);
		\draw [chainLine, ->] (s-1-2) to [out=60,in=120, looseness=0.9] node[above, yshift=-1mm, color=black]{\footnotesize\it prep} (s-1-5);
		\draw [chainLine, ->] (s-1-5) to [out=60,in=120, looseness=1] node[above, yshift=-1mm, color=black]{\footnotesize\it pobj} (s-1-6);
		
		\node[emb, above= of s-1-1, yshift=5mm] (u1) {$\vec{w}_\mathbf{1}$};
		\node[emb, above= of s-1-2, yshift=5mm] (u2) {$\vec{w}_\mathbf{2}$};
		\node[emb, above= of s-1-3, yshift=5.8mm] (u3) {$\vec{w}_\mathbf{3}$};
		\node[emb, above= of s-1-4, yshift=5mm] (u4) {$\vec{w}_\mathbf{4}$};
		\node[emb, above= of s-1-5, yshift=5mm] (u5) {$\vec{w}_\mathbf{5}$};
		\node[emb, above= of s-1-6, yshift=5mm] (u6) {$\vec{w}_\mathbf{6}$};
		
		\node[lstm, above= of u1] (h1) {};
		\node[lstm, above=of u2](h2) {};
		\node[lstm, above= of u3](h3) {};
		\node[lstm, above= of u4](h4) {};
		\node[lstm, above= of u5](h5) {};
		\node[lstm, above= of u6](h6) {};

		\node[gfunc, above= of h1, yshift=-1mm] (g1) {};
		\node[gfunc, above=of h2, yshift=-1mm](g2) {};
		\node[gfunc, above= of h3, yshift=-1mm](g3) {};
		\node[gfunc, above= of h4, yshift=-1mm](g4) {};
		\node[gfunc, above= of h5, yshift=-1mm](g5) {};
		\node[gfunc, above= of h6, yshift=-1mm](g6) {};
		
		%		\node[lstm, above= of g1, yshift=-3mm] (m1) {};
		%		\node[lstm, above=of g2, yshift=-3mm](m2) {};
		%		\node[lstm, above= of g3, yshift=-3mm](m3) {};
		%		\node[lstm, above= of g4, yshift=-3mm](m4) {};
		%		\node[lstm, above= of g5, yshift=-3mm](m5) {};
		%		\node[lstm, above= of g6, yshift=-3mm](m6) {};
		
		%		\node[left = of m1, xshift](layer1) {$L=1$};
		
		\node[sentity, above= of g1] (e1) {\footnotesize \textcolor{fontgray}{\bf \textsc{s-per}}};
		\node[sentity, above=of g2](e2) {\footnotesize \textcolor{fontgray}{\bf \textsc{o}}};
		\node[sentity, above= of g3](e3) {\footnotesize \textcolor{fontgray}{\bf \textsc{o}}};
		\node[sentity, above= of g4](e4) {\footnotesize \textcolor{fontgray}{\bf \textsc{o}}};
		\node[sentity, above= of g5](e5) {\footnotesize \textcolor{fontgray}{\bf \textsc{o}}};
		\node[sentity, above= of g6](e6) {\footnotesize \textcolor{fontgray}{\bf \textsc{s-gpe}}};

		\draw [chainLine] (u1) to (h1);
		\draw [chainLine] (u2) to (h2);
		\draw [chainLine] (u3) to (h3);
		\draw [chainLine] (u4) to (h4);
		\draw [chainLine] (u5) to (h5);
		\draw [chainLine] (u6) to (h6);
		
		\draw [chainLine] (h1) to  (g1);
		\draw [chainLine] (h2) to (g2);
		\draw [chainLine] (h3) to (g3);
		\draw [chainLine] (h4) to (g4);
		\draw [chainLine] (h5) to (g5);
		\draw [chainLine] (h6) to (g6);
		
		%		\draw [chainLine] (g1) to (m1);
		%		\draw [chainLine] (g2) to (m2);
		%		\draw [chainLine] (g3) to (m3);
		%		\draw [chainLine] (g4) to (m4);
		%		\draw [chainLine] (g5) to (m5);
		%		\draw [chainLine] (g6) to (m6);
		
		\draw [chainLine] (g1) to  (e1);
		\draw [chainLine] (g2) to   (e2);
		\draw [chainLine] (g3) to   (e3);
		\draw [chainLine] (g4) to    (e4);
		\draw [chainLine] (g5) to   (e5);
		\draw [chainLine] (g6) to   (e6);
		
		\draw [chainLine, dashed] (h2) to  [out=60,in=-60] (g2);
		\draw [chainLine, dashed] (h2) to (g1);
		\draw [chainLine, dashed] (h2) to (g4);
		\draw [chainLine, dashed] (h2) to (g5);
		\draw [chainLine, dashed] (h4) to (g3);
		\draw [chainLine, dashed] (h5) to (g6);
		
		\draw [chainLine, <->] (h1) to (h2);
		\draw [chainLine, <->] (h2) to (h3);
		\draw [chainLine, <->] (h3) to (h4);
		\draw [chainLine, <->] (h4) to (h5);
		\draw [chainLine, <->] (h5) to (h6);
		
		%		\draw [chainLine, <->] (m1) to (m2);
		%		\draw [chainLine, <->] (m2) to (m3);
		%		\draw [chainLine, <->] (m3) to (m4);
		%		\draw [chainLine, <->] (m4) to (m5);
		%		\draw [chainLine, <->] (m5) to (m6);
		
		\draw [chainLine, line width=1.2pt, -, middlefactor] (e1) to (e2);
		\draw [chainLine, line width=1.2pt, -, middlefactor] (e2)to (e3);
		\draw [chainLine, line width=1.2pt, -, middlefactor] (e3)to (e4);
		\draw [chainLine, line width=1.2pt, -, middlefactor]  (e4)to (e5);
		\draw [chainLine, line width=1.2pt, -, middlefactor] (e5)to (e6);
		
		\end{tikzpicture} 
	}
	\vspace*{-8mm}
	\caption{BiLSTM-GCN-CRF. Dashed connections mimic the dependency edges.}
	%	\vspace*{-5mm}
	\label{fig:gcnmodel}
\end{figure}

Figure \ref{fig:gcnmodel} shows the neural architecture for the BiLSTM-GCN-CRF model. 
Following \citet{zhang2018graph}, the input representation at each position $\vec{w}_i$ is the word representation which consists of the pre-trained word embeddings and its character representation. 
To capture contextual information, we stack a BiLSTM layer before the GCN. 
Secondly, the GCN captures the dependency tree structure as shown in Figure \ref{fig:gcnmodel}. 
Following \citet{zhang2018graph}, we treat the dependency trees as undirected and build a symmetric adjacency matrix during the GCN update:
\begin{equation}
\vec{h}_i^{(l)}
=
\text{ReLU}
\Big(
\sum_{j=1}^{n}
A_{ij}
\mathbf{W}^{(l)} \vec{h}_j^{(l-1)}
+
\vec{b}^{(l)}
\Big)
\label{equ:gcn}
\end{equation}
where $\mathbf{A}$ is the adjacency matrix. $A_{ij} = 1$ indicates there is a dependency edge between the $i$-th word and the $j$-th word\footnote{$A_{ij} = A_{ji}$ for symmetric matrix. }. 
$\vec{h}_i^{(l)}$ is the hidden state at the $i$-th position in the $l$-th layer. 
We can stack $J$ layers of GCN in the model. 
In our experiments, we set the number of GCN layers $J = 1$ as we did not observe significant improvements by increasing $J$. 
In fact, we might obtain harmful performance for a larger $J$ as deeper GCN layers will diminish the effect of the contextual information, which is important for the task of NER. 

However, Equation \ref{equ:gcn} does not include the dependency relation information. 
As mentioned in the main paper, such relations have strong correlations with the entity types. 
We modify the Equation \ref{equ:gcn} and include the dependency relation parameter\footnote{The bias vector is ignore for brevity.}:
\begin{equation*}
\vec{h}_i^{(l)} \!= \sigma \Big(
\! \sum_{j=1}^{n}  \!
A_{ij}
\big ( 
\mathbf{W}_1^{(l)} \vec{h}_j^{(l-1)}
+ \mathbf{W}_2^{(l)}\vec{h}_j^{(l-1) }w_{r_{ij}}
\big )
%+
%\vec{b}^{(l)}
\Big)\nonumber
\end{equation*}
where $w_{r_{ij}}$ is the dependency relation weight that parameterize the dependency relation $r$ between the $i$-th word and the $j$-th word. 
Such formulation uses the relation to weight the adjacent hidden states in the dependencies. 

\section{Implementation Details}
We implemented all the models with PyTorch~\cite{paszke2017automatic}.
\begin{figure*}[t!]
	\centering
	\includegraphics[width=6.2in]{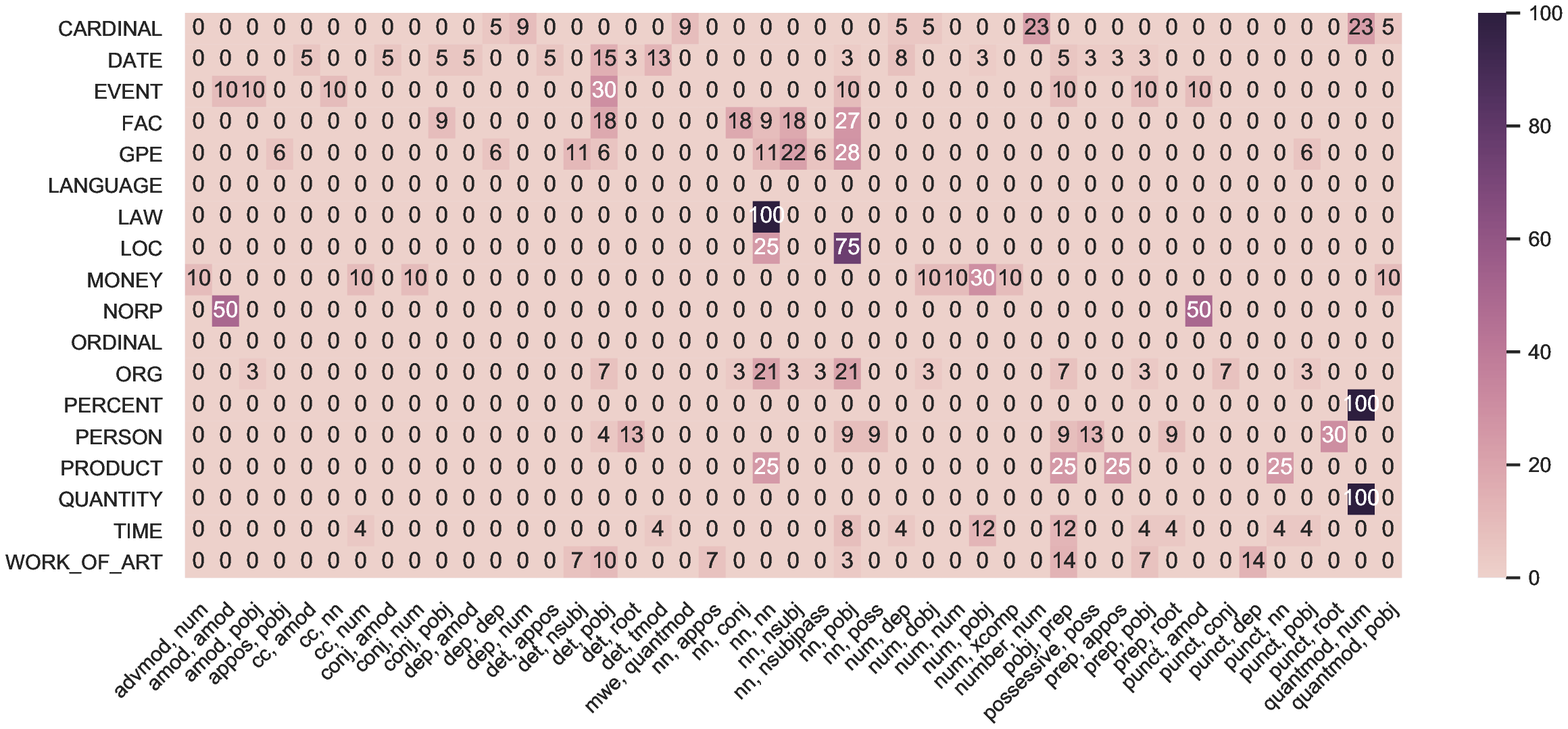}
	\caption{Correlations between the entity types and the dependency relation pairs on the grandchild dependencies.}
	\label{fig:gd}
\end{figure*}
For both BiLSTM-CRF and DGLSTM-CRF model, we train them on all datasets with 100 epochs and take the model that perform the best on the development set. 
For BiLSTM-GCN-CRF, we train for 300 epochs with a clipping rate of 3.

\section{Relation Pairs on Grandchild Dependencies}
Figure \ref{fig:gd} visualized the correlations between the entities and the grandchild dependency relation pairs on the OntoNotes English dataset. 
As mentioned in the paper, such entities are correctly predicted by our models but not the BiLSTM-CRF baseline. 
As we can see from the figure, most of these entities correlate to the ``(\textit{nn}, \textit{nn})'' and ``(\textit{nn}, \textit{pobj})'' relation pairs on the grandchild dependencies. 
Such correlations also show that the relation pair information on the grandchild dependencies can be helpful for detecting certain entities.

\section{Using Predicted Dependency}

We train a BERT-based~\cite{devlin2019bert} dependency parser~\cite{dozat2017deep} using the training set for each of four languages. 
Specifically, we adopt the \texttt{bert-base-uncased} model for English, \texttt{bert-base-multilingual-cased} for Catalan and Spanish and \texttt{bert-base-chinese} for Chinese.
Because the Chinese BERT model is based on characters but not Chinese words which are segmented. 
We further incorporate a span extractor layer right after BERT encoder for Chinese. 
We following \citet{lee2017end} to design the span extractor layer. Our code for dependency parser is available at \url{https://github.com/allanj/bidaf_dependency_parsing}

\end{document}